\documentclass{IEEEtran}

\usepackage{graphicx}
\usepackage{epstopdf}
\usepackage{amsmath,amssymb,booktabs,tabularx}
\usepackage[hyphens]{url}
\usepackage{hyperref}
\usepackage{mathrsfs}
\usepackage{float}
\usepackage{multirow}
\usepackage{soul}
\usepackage[table]{xcolor}
\usepackage{comment}
\usepackage{subcaption}
\usepackage{mwe}
\usepackage{slashbox}
\usepackage{authblk}
\usepackage{multicol}
\usepackage{import}
\usepackage{siunitx}
\usepackage{mathtools}
\usepackage{standalone}
\usepackage{makecell} 

\newcommand{\etal}[1]{#1~\emph{et~al.}}

\title{GMM-Based Synthetic Samples for Classification of Hyperspectral Images With Limited Training Data}
\author{AmirAbbas Davari, Erchan Aptoula, Berrin Yanikoglu, Andreas Maier, Christian Riess
\thanks{A. Davari, A. Maier and C. Riess are with Friedrich-Alexander University Erlangen-Nuernberg, 91058 Erlangen, Germany (email: amir.davari@fau.de)}
\thanks{E. Aptoula is with Gebze Technical University, 41400 Gebze, Turkey}
\thanks{B. Yanikoglu is with Sabanci University, 34956 Istanbul, Turkey}}

\begin{document}
\maketitle
\begin{abstract}
The amount of training data that is required to train a classifier scales with
the dimensionality of the feature data. In hyperspectral remote sensing,
feature data can potentially become very high dimensional. However, the amount
of training data is oftentimes limited. Thus, one of the core challenges in
hyperspectral remote sensing is how to perform multi-class classification using only relatively few training data points.

In this work, we address this issue by enriching the feature matrix with 
synthetically generated sample points. This synthetic data is sampled from a GMM fitted to each class of the limited training data. Although, the true distribution
of features may not be perfectly modeled by the fitted GMM, we demonstrate that a moderate augmentation by these synthetic samples can effectively replace a part of the missing training
samples. We show the efficacy of the proposed approach on two hyperspectral
datasets. The median gain in classification performance is $5\%$. It is also encouraging that this performance gain is remarkably stable for large
variations in the number of added samples, which makes it much easier to apply
this method to real-world applications.
\end{abstract}



\section{Introduction}\label{introduction}

\IEEEPARstart{R}{emote} sensing is undoubtedly of paramount importance for several
application fields, including environmental monitoring, urban planning,
ecosystem-oriented natural resources management, urban change detection and
agricultural region monitoring \cite{valero2013hyperspectral}. 
In particular, hyperspectral remote sensing (HSRS) makes use of data with a
spectral resolution that is considerably higher than what off-the-shelf color
cameras provide. The task of HSRS classification is the construction of a label
map of remotely sensed images in which individual pixels are marked as members
of specific classes like water, asphalt, or grass. The decision for the region
type that is seen in a pixel is typically made by a classifier.

Labeling the remote-sensing data is typically a manual, expensive and
time-consuming process, involving pixel-level details. Thus, it is very common that many publicly available datasets contain ground-truth
labels for only a small subset of pixels for each of the (potentially many)
classes.

Current work that addresses the limited training data HSRS image classification can be
roughly divided into two categories. In the first category the aim is to develop classifiers that are more robust to limited training data, e.g.,~\cite{hoffbeck1996covariance,tadjudin1998covariance,bruzzone2006novel,chi2008classification,xia2016rotation,li2015multiple}. 
In the second category,
the aim is to reduce the feature dimensionality since the limited data problem is less severe in lower-dimensional spaces,
e.g.~\cite{sofolahan2013summed,kuo2004nonparametric,fukunaga2013introduction,lee1993feature,castaings2010influence,kianisarkaleh2016nonparametric}.
While dimensionality reduction methods have been proven to be useful in many different problems, current methods are highly challenged in extreme cases, i.e., when
training data is severely limited. For example, when reducing the number of
training samples per class from $40$ to $13$, the average accuracy for a
standard pipeline that computes PCA, then extended multi-attribute profiles (EMAP) features, then PCA again drops
on the Pavia Centre dataset from about $84\%$ to about $74\%$. 


Probably any machine learning method can benefit from augmentation techniques. Speaking of which one could also try to model more of the imaging process for the augmentations, e.g. signal generation, noise, etc. However, this requires a very accurate model in order to work. If the model is not good enough, transfer learning techniques \cite{heimann2014real} yield significant improvements. If it is possible to carry out better simulation, it may be possible to get away without transfer learning and successive labeling \cite{unberath2017respiratory}.
We observed that it is possible to adapt the data to the classifier, by augmenting the data with synthetic samples.
We noted this option earlier in a conference
paper~\cite{davari2015effect}, but a thorough examination is still missing. This
approach to limited data classification is discussed, and quantitatively and
qualitatively analyzed. It is observed that adding synthetic data yields
excellent results at a very low computational cost.




This paper is organized as follows. In Sec.~\ref{related_work}, we briefly
review the related work on the remote-sensing limited-data
classification. The explored approach is detailed in
Sec.~\ref{methodology}. Next, Sec.~\ref{Experimental_setup} presents the
conducted experiments and their results, and puts them in perspective with
other works. Finally, Sec.~\ref{Conclusion} concludes the paper.

\section{Related Work}\label{related_work}

We organize the related methods into two groups, namely robust classification schemes and 
dimensionality reduction methods. 
It is worth mentioning that in the context of synthetic data generation, there exist many works based on generative adversarial networks (GANs) \cite{goodfellow2014generative} such as generating realistic synthetic training data \cite{shrivastava2016learning}. These deep neural network-based methods are extremely data demanding and are very unusual to be applied on severely limited data, e.g. few pixels.

\subsection{Robust Classification}
Early works used a Gaussian maximum likelihood
estimator~\cite{hoffbeck1996covariance,tadjudin1998covariance}. However,
limited training data leads to inaccuracies in the estimation of the Gaussian
means and covariances. This was addressed by modifying the covariance matrix
estimation.


\etal{Bruzzone} proposed to introduce transductive and inductive functions as controlling units on the SVM outputs to select semi-labeled training data~\cite{bruzzone2006novel}. \etal{Chi} modified a support vector machine (SVM), and performed gradient descent and a Newton-Raphson optimization on its primal representation~\cite{chi2008classification}.

Recently, \etal{Xia} proposed the rotation based SVM (RoSVM), which is a novel SVM-based ensemble approach \cite{xia2016rotation}. They use random feature selection in order to diversify the classifier's result. Their experiments show an enhanced performance on the limited training data, compared to normal SVM. However, their method is computationally expensive.

\etal{Li} proposed a framework for hyperspectral remote
sensing image classification which is based on integrating multiple linear and
non-linear features, including EMAP~\cite{li2015multiple}. In contrast to
kernel methods which is specified to either linear or non-linear features,
their framework can take care of both linear and non-linear features and
integrate them into a more effective classifier.


%

\subsection{Dimensionality Reduction}
Dimensionality reduction (DR) algorithms are widely exploited in hyperspectral
remote sensing image classification to reduce the number of spectral channels.
They directly address the Hughes phenomenon by reducing the dimensionality of
the feature vector. Principle component analysis (PCA) and independent component
analysis (ICA) are two of the most commonly used DR algorithms in the
literature.

\etal{Sofolahan} proposed an algorithm named summed component analysis that
uses PCA and principle feature analysis (PFA)~\cite{sofolahan2013summed}. PFA
selects a subset of the features, which in contrast to PCA and ICA allows to
physically interpret the reduced features.
Supervised DR has the added benefit of making use of the
labeled data during DR. Some of the most popular
approaches are non-parametric weighted feature extraction (NWFE)
\cite{kuo2004nonparametric}, discriminant analysis feature extraction (DAFE)
\cite{fukunaga2013introduction}, and decision boundary feature extraction
(DBFE) \cite{lee1993feature}. These approaches were shown to perform equally
well or better than unsupervised reduction techniques, and to boost
classification performance when used in combination with the unsupervised
techniques~\cite{castaings2010influence}. The common idea behind all these
algorithms is to map the data to another space, calculate the scatter matrix
and minimize the within-class and maximize the between-class overall distance
in lower dimension.

Recently, \etal{Kianisarkaleh} proposed nonparametric feature extraction (NFE) as a new way of DR~\cite{kianisarkaleh2016nonparametric}. These features exhibit improved performance when dealing with limited training set population. Its idea is very similar to NWFE, however it uses $k$ neighboring samples in a class $i$ to compute the local class mean.

%
%
%



\section{Addition of Synthetic Data for Classification}\label{methodology}

\begin{figure}[tb]
	\centering
	\centerline{\includegraphics[width=1\linewidth]{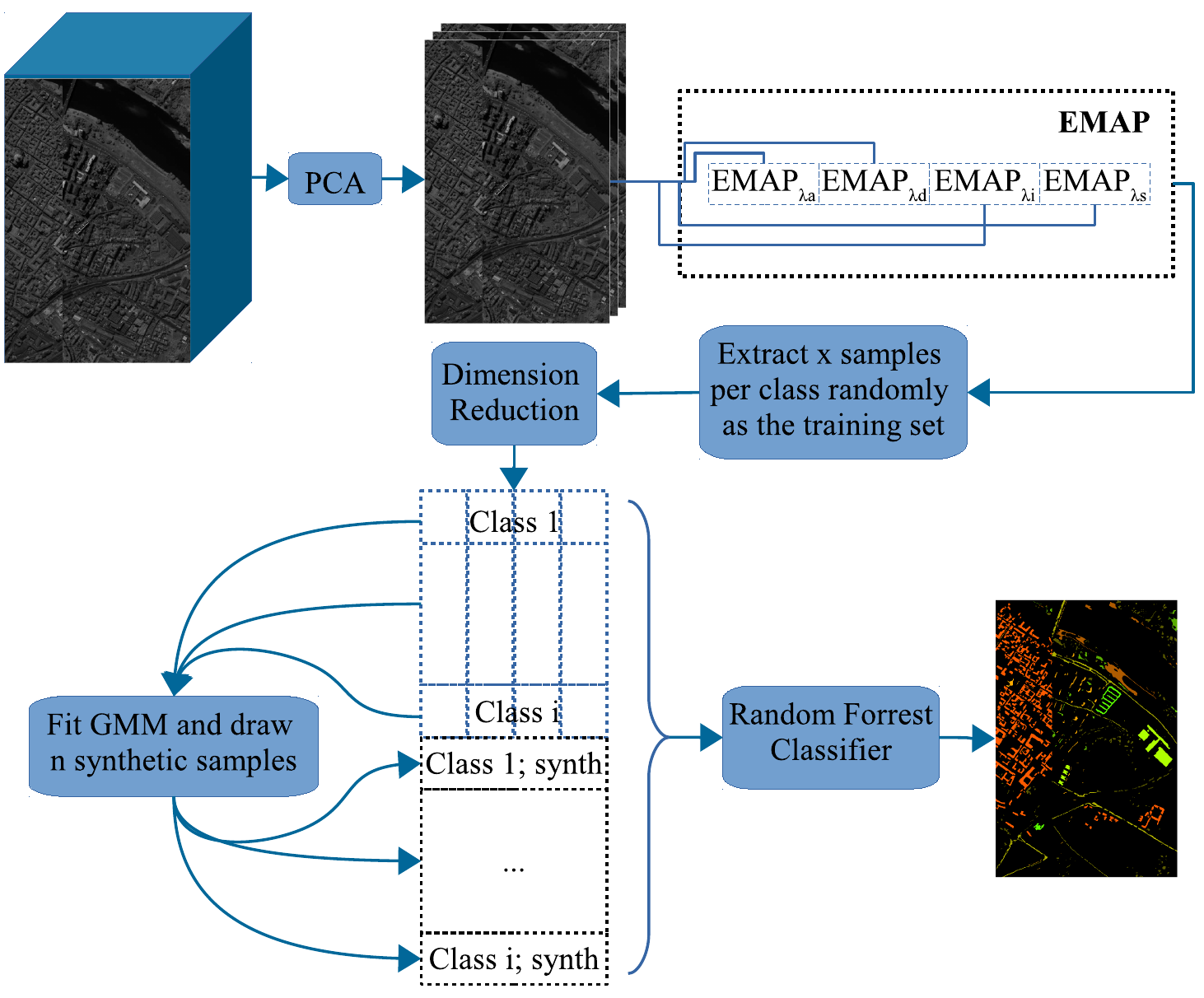}}
	\caption{Proposed workflow.}
	\label{workflow}
\end{figure}

A high-level overview of the proposed method is shown in Fig.~\ref{workflow}. We use a standard dimensionality-reduction workflow where spectral bands are first reduced via PCA. Then, extended multi-attribute profiles (EMAP)~\cite{dalla2010extended} are computed as a feature vector. These features are again subject to dimensionality reduction. These low-dimensional descriptors are then fed into the classifier. The key contribution of the method is injected right before the classification: we propose to populate the feature space more densely with synthetic feature points. These feature points are drawn from a Gaussian Mixture Model (GMM) that is fitted to the actual (few) training samples. A GMM from such limited training data is necessarily only a coarse approximation of the underlying distribution. Nevertheless, we show that it is good enough to support the classifier in better determining the class boundaries.

The parametrization of the standard pipeline follows dataset-dependent
recommendations from the literature, and is reported in the experiments in
Sec.~\ref{Experimental_setup}. For the remainder of this section, we expand on
our core contribution, which is the addition of synthetic data.

A GMM models the probability density function (PDF) as
\begin{equation}
	p(\boldsymbol{x}) = \sum\limits_{i=1}^{k} w_i \mathcal{N}(\boldsymbol{x}|\boldsymbol{\mu}_i,\boldsymbol{\Sigma}_i) \quad \mathrm{s.t.} \sum\limits_{i=1}^k w_i = 1\enspace,
\end{equation}
where $\boldsymbol{x} \in \mathbb{R}^d$ denotes a $d$-dimensional sample, $k$
is the number of mixture components, $w_i$, $0\le w_i \le 1$, is the weight of the $i$-th component,
and $\mathcal{N}(\boldsymbol{x}|\boldsymbol{\mu}_i,\boldsymbol{\Sigma}_i)$ is
the a posteriori probability of $\boldsymbol{x}$ given the multivariate
Gaussian distribution with mean vector $\boldsymbol{\mu}_i$ and covariance
$\boldsymbol{\Sigma}_i$.

The Gaussian mixture model is fully
parameterized by the coefficients $w_i$, the mean vectors $\boldsymbol{\mu}_i$
and the covariance matrices $\boldsymbol{\Sigma}_i$. Thus, the total number of
parameters is $k+kd+kd^2$ for $k$ components of dimension $d$.  
When facing severely limited data,
there may not be enough samples available for accurately parameterizing the full model. As a consequence, we
constrain the covariance matrices to diagonal matrices. Such a linear
combination of diagonal matrices is sufficient to model correlation between
dimensions~\cite{reynolds2009gaussian}. Thus, the benefit of a full covariance
matrix can be assumed to be minor compared to the fact that the number of
estimated parameters greatly decreases to only $k+kd+kd = k(1+2d)$.

To allow for some variability in the number of components of each GMM model, we construct for each class four GMMs with $k = 1$ to $k=4$ components. The best fitting model is determined using the Akaike information  criterion~\cite{mclachlan2004finite,oliveira2005assessing}.

GMM parameter estimation is being carried out by iterative {\color{black}{EM (MAP)}},
which is quite sensitive to the initial values. Thus, we use the k-means
clustering algorithm to provide reasonable initial values for the estimator,
where $k$ is set to the selected number of components.
Further, by adding a small value to the diagonal of the covariance matrices, it
is ensured that EM will not get stuck in an ill condition and will converge.

\section{Evaluation}\label{Experimental_setup}

We use the popular Pavia Centre and Salinas datasets for evaluation. The Pavia
Centre dataset has been acquired by the ROSIS sensor in $115$ spectral bands
over Pavia, northern Italy. $13$ of these bands are removed due to noise and
therefore $102$ bands are used in this work. The scene image is $1096\times715$
pixels with a geometrical resolution of $1.3\,\mathrm{m}$. Salinas dataset was
acquired by AVIRIS sensor in 224 spectral bands over Salinas Valley,
California. $20$ water absorption bands were discarded and the remaining $204$
bands are used in this work. The image is $512\times 217$ pixels with a
geometrical resolution of $3.7\,\mathrm{m}$. 

Fig.~\ref{workflow} shows our standard hyperspectral remote sensing
classification pipeline that is based on dimensionality reduction.
First, PCA is performed on the input data to preserve $99\%$ of the total
spectral variance. On these PCA components, extended multi-attribute profile
(EMAP) features are computed. We followed the literature by using
four attributes and four thresholds $\lambda$ per
attribute~\cite{dalla2010extended,liu2016class}. 
More specifically, the thresholds for area of
connected components are chosen as $\lambda_a = [100, 500, 1000,
5000]$, and the thresholds for length of the diagonal of the bounding box fitted around the connected components 
$\lambda_d$ are chosen as $\lambda_d = [10, 25, 50, 100]$. 
The thresholds for standard deviation of the gray values of the connected components
$\lambda_s$ and the moment of inertia $\lambda_i$ are chosen differently per
dataset~\cite{dalla2010extended,liu2016class}, i.e., $\lambda_s = [20, 30, 40,
50]$ and $\lambda_i = [0.2, 0.3, 0.4, 0.5]$ for Pavia Centre, and for Salinas
$\lambda_s = [20, 30, 40, 50]$ and $\lambda_i = [0.1, 0.15, 0.2, 0.25]$.
It is worth noting that our main focus in this paper is bringing the proposed method's effectiveness under study and not achieving the highest classification performance. That, along with the popularity of PCA and the aforementioned threshold values in the literature, motivate us to use them in our evaluation. However, we acknowledge that neither are optimal.

For the second dimensionality reduction, we use
in one variant the unsupervised PCA, and in another variant the
supervised non-parametric weighted feature extraction
(NWFE)~\cite{kuo2004nonparametric,castaings2010influence} to preserve $99\%$ of
the feature variance. On Pavia Centre, PCA and NWFE result in 7 and 6 feature
dimensions, respectively. On Salinas, PCA and NWFE result in 4 and 7
dimensions, respectively. In our experiments, we use abbreviations to specify
the used pipeline configuration. We use either EMAP, EMAP-PCA, or EMAP-NWFE to
distinguish the use of no secondary dimensionality reduction, PCA, or NWFE,
respectively.
Classification is performed with random forest. 
Each experiment is repeated $25$ times and the mean average accuracy (AA), overall accuracy (OA) and Kappa along with their standard deviations are reported.

\begin{figure}[tb]
	\begin{minipage}[b]{1\linewidth}
		\centering
		\centerline{\includegraphics[width=1\linewidth]{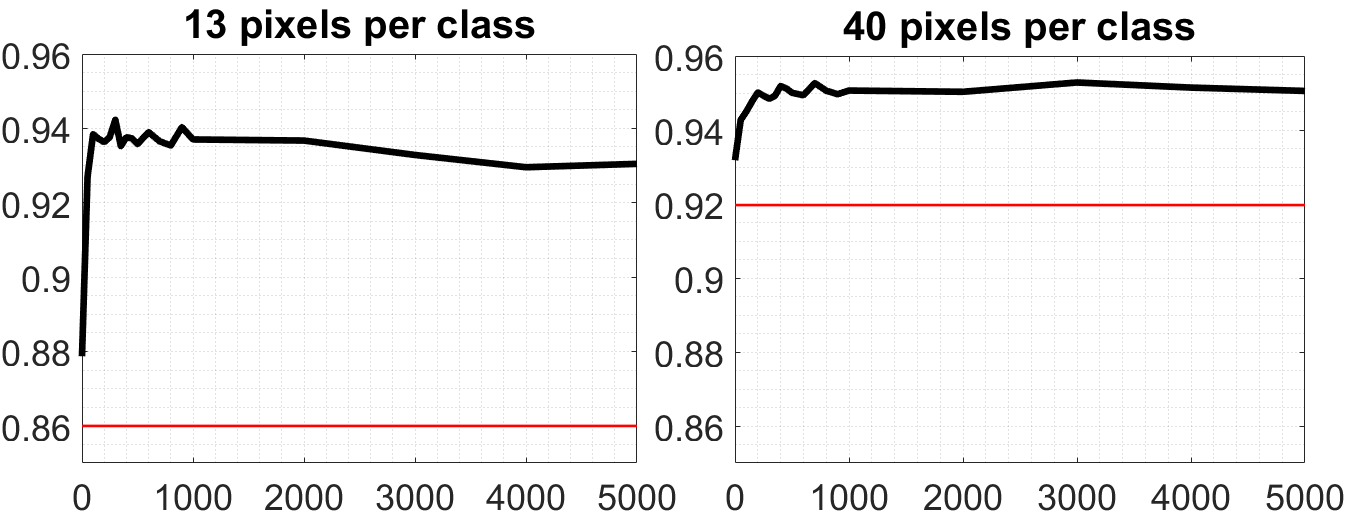}}
	\end{minipage}
	\caption{Classification performance (kappa) vs. the number of synthetic samples added to the original training set. Classification is performed on EMAP-NWFE computed over Pavia Centre dataset. Red line represents the performance of raw EMAP without any synthetic sample addition.}
	\label{fig:result_num_samples}
\end{figure}

A first result is shown in Fig.~\ref{fig:result_num_samples}. Here, we
performed classification on 13 (left) and 40 (right) training samples per
class, respectively, on Pavia Centre dataset. 
We use the random forest default parameters as proposed by Breiman~\cite{breiman2001random}, i.e. $H=100$ trees with a tree depth of the square root of the feature dimension, $D=\sqrt{d}$, on EMAP-NWFE, and report Kappa for different numbers of up to $5000$ added synthetic samples. It turns out that adding only a few synthetic samples leads to a jump in classification performance, e.g. from about $0.88$ to about $0.94$ if 13 training samples per class are used. This performance gain is quite stable with respect to the exact number of added samples, i.e., it does not make much difference whether 500 or 5000 samples are added.

A full quantitative evaluation is performed on EMAP, EMAP-PCA, and EMAP-NWFE which are computed on Pavia Centre and Salinas dataset, using random forests classifier.
Since we require a low dimensional data in order our GMM parameter estimation to converge with the limited available data, synthetic samples are
added to the dimensionality-reduced EMAP-PCA and EMAP-NWFE, but not to the
high-dimensional EMAP space.
Representative example results are shown in Tab.~\ref{tab:rf_synth_results}. The class-wise performances for Pavia Centre and Salinas datasets are given in Tab.~\ref{tab:pavia_classwise} and Tab.~\ref{tab:salinas_classwise}. 
In every case, the variants using synthetic samples improve the classification
performance. We confirmed these findings also
for 20 and 30 training samples per class, and we also repeated and confirmed
all experiments with SVM classifier (see supplemental material). The average
improvement of kappa jumps $5.84$ percentile points up after adding synthetic
samples to the training set, with a standard deviation of $3.18$. 

These results are also competitive with other methods reported in the literature. 
The results reported by \etal{Kianisarkaleh}~\cite{kianisarkaleh2016nonparametric}, computed using limited training samples on the Salinas dataset are very comparable to ours.
\etal{Li}~\cite{li2015multiple} also recently proposed a framework to operate on very limited datasets. The overall accuracy (OA) of their results reported on Pavia Centre dataset for $20$, $30$, and $40$ training samples is slightly higher than our OA. However, their reported average accuracy is comparable to our performance, and the kappa values of our approach are in all considerably higher. 
\etal{Aptoula}~\cite{aptoula2016deep} use deep learning for classification. Their kappa on Pavia Centre for the full spectral dataset is $0.952$, which is very close to the random forest performance EMAP-NWFE-Synth kappa of $0.9528$ for $40$ samples. Deep learning on area and moment attribute profiles yielded a best-case kappa of $0.983$, which is better than our results. However, in all cases, their methods operate on a training set that is, depending on the class, six to 14 times larger than ours. 
\etal{Tao} use a deep autoencoder to learn the features for hyperspectral image classification~\cite{tao2015unsupervised}. In a feature transfer task, they report on Pavia Centre a kappa of $0.9699$ using $50$ samples from the dataset, which is somewhat higher than our EMAP-NWFE-Synth on $40$ samples ($0.9528$). However, key to their strong performance is to first learn a sophisticated feature representation from the Pavia University dataset using a considerably higher number of samples. In future work, it would be interesting to investigate whether their feature representation can also benefit from additional synthetic samples. 
Furthermore, we run our pipeline on Pavia University dataset in order to quantitatively compare our work with a recent work on limited training data by \etal{Xia}~\cite{xia2016rotation}. Our $[OA, AA] = [79.52, 84.78]$ on $13$ pixels per class training set size, computed on the EMAP-PCA is higher than their best result using $30$ pixels per class training set size, i.e. $[76.06, 82.67]$ and almost equal to their result on $40$ pixels per class training set, i.e. $[77.12, 85.12]$ (see supplemental material for the full results).
All in all, it is encouraging that our proposed approach is able to achieve a performance that comes close to a deep learning architecture, which may be very useful in scenarios where there is not the significant amount of training data available that is required to train a deep network.


Two observations can be made from the results in Tab.~\ref{tab:rf_synth_results}. First, the addition of synthetic samples not only outperforms the EMAP-PCA and EMAP-NWFE, but also it results in higher performance than the raw non-reduced EMAP. Second, adding synthetic samples reduces the standard deviation of the classification results. In other words, populating the training data with the synthetic samples helps the classifier in reducing the uncertainty when being fed by different limited and randomly selected training data. Both observations indicate that the added synthetic samples which are generated by our proposal are well simulating and representing the hyperspectral images under study.

\begin{table}[tb]
	\centering
	\caption{Classification performance computed over Pavia Centre and Salinas.
		$|S|$ denotes the number of added synthetic samples per class, ``-'' indicates that no samples are added.}
	\begin{tabular}{|l|c@{ }|c@{ }c@{ }c|}
		\hline
		Algorithm & $|S|$ & AA\% ($\pm$SD) & OA\% ($\pm$SD) & Kappa ($\pm$SD) \\
		\hline
		\hline
		\rowcolor[rgb]{ .706,  .702,  .702} \multicolumn{5}{|c|}{Pavia Centre} \\
		\hline
		\rowcolor[rgb]{ .906,  .902,  .902} \multicolumn{5}{|c|}{13 pix/class} \\
		EMAP & - & 77.87 ($\pm$2.97) & 90.01 ($\pm$3.78) & 0.8600 ($\pm$0.0495)  \\
		\hline
		EMAP-PCA  & -   & 73.51 ($\pm$3.00) & 86.38 ($\pm$3.61) & 0.8089 ($\pm$0.0493)  \\
		EMAP-PCA  & 500 & \textbf{84.59} ($\pm$1.58) & \textbf{93.67} ($\pm$0.75) & \textbf{0.9107} ($\pm$0.0104)  \\
		\hline
		EMAP-NWFE & -   & 80.06 ($\pm$3.56) & 91.37 ($\pm$2.67) & 0.8787 ($\pm$0.0365)\\
		EMAP-NWFE & 500 & \textbf{89.57} ($\pm$1.15)  & \textbf{95.91} ($\pm$0.49)  & \textbf{0.9423} ($\pm$0.0069) \\
		\hline
		\rowcolor[rgb]{ .906,  .902,  .902} \multicolumn{5}{|c|}{40 pix/class} \\
		
		EMAP & - & 86.80 ($\pm$1.47) & 94.30 ($\pm$0.61) & 0.9197 ($\pm$0.0086)  \\
		\hline
		EMAP-PCA  & -   & 83.98 ($\pm$1.18) & 93.49 ($\pm$0.69) & 0.9082 ($\pm$0.0096)  \\
		EMAP-PCA  & 500 & \textbf{88.74} ($\pm$0.96) & \textbf{95.09} ($\pm$0.54) & \textbf{0.9307} ($\pm$0.0076)  \\
		\hline
		EMAP-NWFE & -   & 87.41 ($\pm$1.41)  & 95.18 ($\pm$0.61)  & 0.9318 ($\pm$0.0085) \\
		EMAP-NWFE & 500 & \textbf{92.39} ($\pm$0.75) & \textbf{96.66} ($\pm$0.46) & \textbf{0.9528} ($\pm$0.0063)\\
		\hline
		\hline
		\rowcolor[rgb]{ .706,  .702,  .702} \multicolumn{5}{|c|}{Salinas} \\
		\hline
		\rowcolor[rgb]{ .906,  .902,  .902} \multicolumn{5}{|c|}{13 pix/class} \\
		EMAP    & - & 83.84 ($\pm$2.06) & 76.30 ($\pm$2.74) & 0.7380 ($\pm$0.0292)  \\
		\hline
		EMAP-PCA  & -   & 82.50 ($\pm$2.06) & 74.96 ($\pm$3.63) & 0.7230 ($\pm$0.0378)  \\
		EMAP-PCA  & 500 & \textbf{90.96} ($\pm$0.88) & \textbf{83.89} ($\pm$1.72) & \textbf{0.8215} ($\pm$0.0188)  \\
		\hline
		EMAP-NWFE & -   & 88.68 ($\pm$1.20)  & 80.42 ($\pm$2.34)  & 0.7838 ($\pm$0.0247)  \\
		EMAP-NWFE & 500 & \textbf{93.17} ($\pm$0.68) & \textbf{87.09} ($\pm$1.26) & \textbf{0.8566} ($\pm$0.0138)  \\
		\hline
		\rowcolor[rgb]{ .906,  .902,  .902} \multicolumn{5}{|c|}{40 pix/class} \\
		EMAP    & - & 90.75 ($\pm$0.86) & 84.52 ($\pm$1.76) & 0.8285 ($\pm$0.0192)  \\
		\hline
		EMAP-PCA  & -   & 89.80 ($\pm$1.21) & 81.73 ($\pm$2.51) & 0.7981 ($\pm$0.0273)  \\
		EMAP-PCA  & 500 & \textbf{93.08} ($\pm$0.40) & \textbf{86.59} ($\pm$0.85) & \textbf{0.8512} ($\pm$0.0093)  \\
		\hline
		EMAP-NWFE & -   & 93.29 ($\pm$0.41)  & 86.09 ($\pm$1.86)  & 0.8462 ($\pm$0.0200)  \\
		EMAP-NWFE & 500 & \textbf{94.52} ($\pm$0.30) & \textbf{89.18} ($\pm$0.81) & \textbf{0.8798} ($\pm$0.0089)  \\
		\hline
		
	\end{tabular}%
	\label{tab:rf_synth_results}%
\end{table}%

\begin{table}[htbp]
	\centering
	\caption{Class-wise performance computed over Pavia Centre dataset, with 500 synthetic samples per class. }
	\begin{tabular}{|l@{ }|@{ }c@{ }||c@{ }|@{ }c@{ }|}
		\hline
		Class & Train/Test & EMAP-PCA & EMAP-NWFE \\
		\hline
		\hline
		
		Water & 13/65958 & 99.48 $\pm$ 0.37 & 99.71 $\pm$ 0.28 \\
		Trees & 13/7585 & 72.09 $\pm$ 10.12 & 83.86 $\pm$ 7.19 \\
		Asphalt & 13/3077 & 68.25 $\pm$ 9.92 & 82.25 $\pm$ 9.65 \\
		Self-Blocking Bricks & 13/2672 & 72.08 $\pm$ 9.82 & 81.80 $\pm$ 6.29 \\
		Bitumen & 13/6571 & 80.47 $\pm$ 7.48 & 80.84 $\pm$ 7.68 \\
		Tiles & 13/9235 & 92.63 $\pm$ 4.17 & 96.39 $\pm$ 1.83 \\
		Shadows & 13/7274 & 84.12 $\pm$ 5.49 & 86.01 $\pm$ 4.93 \\
		Meadows & 13/42813 & 95.25 $\pm$ 2.04 & 98.19 $\pm$ 0.96 \\
		Bare Soil & 13/2850 & 96.97 $\pm$ 2.06 & 97.08 $\pm$ 2.02 \\
		\hline
		\hline
		\multicolumn{2}{|c||}{Average Accuracy} & 84.59 $\pm$ 1.58 & 89.57 $\pm$ 1.15 \\
		\multicolumn{2}{|c||}{Overall Accuracy} & 93.67 $\pm$ 0.75 & 95.91 $\pm$ 0.49 \\
		\multicolumn{2}{|c||}{Kappa} & 0.9107 $\pm$ 0.0104 & 0.9423 $\pm$ 0.0069 \\
		\hline
	\end{tabular}%
	\label{tab:pavia_classwise}%
\end{table}%

\begin{table}[htbp]
	\centering
	\caption{Class-wise performance computed over Salinas dataset, with 500 synthetic samples per class.}
	\begin{tabular}{|l@{ }|@{ }c@{ }||c@{ }|@{ }c@{ }|}
		\hline
		Class & Train/Test & EMAP-PCA & EMAP-NWFE \\
		\hline
		\hline
		Brocoli green weeds 1 & 13/1996 & 95.90 $\pm$ 5.47 & 98.05 $\pm$ 3.65 \\
		Brocoli green weeds 2 & 13/3713 & 95.57 $\pm$ 2.77 & 96.81 $\pm$ 3.98 \\
		Fallow & 13/1963 & 88.82 $\pm$ 6.68 & 97.28 $\pm$ 3.35 \\
		Fallow rough plow & 13/1381 & 99.42 $\pm$ 0.55 & 98.81 $\pm$ 1.65 \\
		Fallow smooth & 13/2665 & 96.61 $\pm$ 1.15 & 94.95 $\pm$ 1.84 \\
		Stubble & 13/3946 & 96.15 $\pm$ 2.04 & 98.24 $\pm$ 1.38 \\
		Celery & 13/3566 & 99.41 $\pm$ 0.22 & 99.75 $\pm$ 0.05 \\
		Grapes untrained & 13/11258 & 58.97 $\pm$ 8.98 & 62.10 $\pm$ 9.88 \\
		Soil vinyard develop & 13/6190 & 95.79 $\pm$ 1.35 & 98.61 $\pm$ 0.86 \\
		Corn
		& 13/3265 & 84.72 $\pm$ 4.90 & 89.49 $\pm$ 6.25 \\
		Lettuce romaine
		4wk & 13/1055 & 92.36 $\pm$ 4.35 & 95.11 $\pm$ 1.65 \\
		Lettuce romaine 5wk & 13/1914 & 94.89 $\pm$ 4.78 & 99.52 $\pm$ 0.57 \\
		Lettuce romaine 6wk & 13/903 & 97.85 $\pm$ 0.95 & 98.66 $\pm$ 0.65 \\
		Lettuce romaine 7wk & 13/1057 & 91.65 $\pm$ 4.18 & 94.62 $\pm$ 2.08 \\
		Vinyard untrained & 13/7255 & 67.55 $\pm$ 6.64 & 68.76 $\pm$ 8.80 \\
		Vinyard vertical trellis & 13/1794 & 99.74 $\pm$ 0.35 & 99.97 $\pm$ 0.11 \\
		\hline
		\hline
		\multicolumn{2}{|c||}{Average Accuracy} & 90.96 $\pm$ 0.88 & 93.17 $\pm$ 0.68 \\
		\multicolumn{2}{|c||}{Overall Accuracy} & 83.89 $\pm$ 1.72 & 87.09 $\pm$ 1.26 \\
		\multicolumn{2}{|c||}{Kappa} & 0.8215 $\pm$ 0.0188 & 0.8566 $\pm$ 0.0138 \\
		\hline
	\end{tabular}%
	\label{tab:salinas_classwise}%
\end{table}%

\begin{figure}[tb]
	
	\begin{minipage}[b]{0.19\linewidth}
		\centering
		\centerline{\includegraphics[width=1\linewidth]{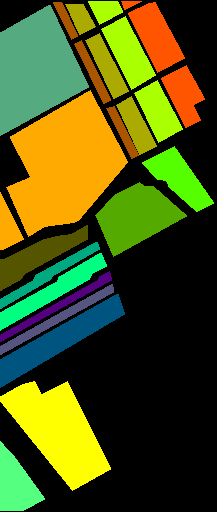}}
		\centerline{(a)}\medskip
	\end{minipage}
	\begin{minipage}[b]{0.19\linewidth}
		\centering
		\centerline{\includegraphics[width=1\linewidth]{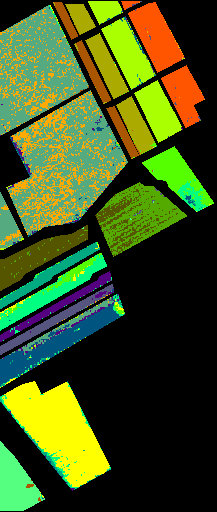}}
		\centerline{(b)}\medskip
	\end{minipage}
	\begin{minipage}[b]{0.19\linewidth}
		\centering
		\centerline{\includegraphics[width=1\linewidth]{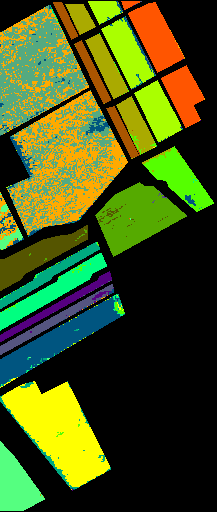}}
		\centerline{(c)}\medskip
	\end{minipage}
	\begin{minipage}[b]{0.19\linewidth}
		\centering
		\centerline{\includegraphics[width=1\linewidth]{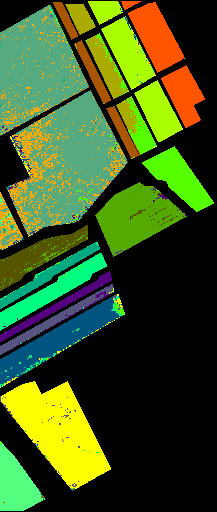}}
		\centerline{(d)}\medskip
	\end{minipage}
	\begin{minipage}[b]{0.19\linewidth}
		\centering
		\centerline{\includegraphics[width=1\linewidth]{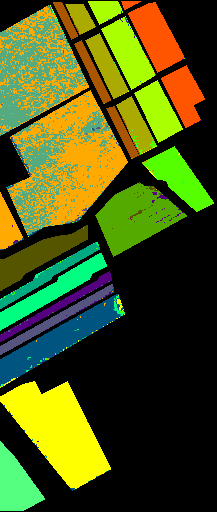}}
		\centerline{(e)}\medskip
	\end{minipage}
	\caption{Label maps on Salinas using $13$ training samples per
		class. (a) ground truth (b) EMAP-PCA, (c) EMAP-PCA with 500 synthetic samples per class, (d) EMAP-NWFE, (e) EMAP-NWFE with 500 synthetic samples per class.}
	\label{Salinas_LM}
\end{figure}

Fig.~\ref{Salinas_LM} shows the selected random forest label maps on Salinas dataset variations with and without adding synthetic samples. The synthetic samples improve the classification accuracy and avoid some misclassification. This improvement can best be observed in the large homogeneous regions.

Our MATLAB implementation is executed on a desktop PC with a quad-core Intel Core i7-4910MQ CPU with 2.9\,GHz and 32\,GB RAM. We report the runtime for generating, adding and classifying $5000$ synthetic samples. It turns out that our method is computationally cheap. For example, for EMAP-PCA on Pavia Centre, it takes \SI{0.68}{\second} to generate, add and classify the synthetic samples. For EMAP-PCA on Salinas dataset, the process takes \SI{1.1}{\second}.

\section{Conclusion}\label{Conclusion}


A common issue in hyperspectral remote sensing image classification is the limited
training data. This limitation severely challenges classifiers, particularly
when using high dimensional feature vectors. In this work, we propose to
compensate this limitation by adding synthetic samples drawn
from a Gaussian mixture that is estimated on the feature space.

We show on the simulated data with non-Gaussian distributions that this idea indeed
helps on severely limited training data, even if the true underlying
distribution is only approximately matched (see supplemental material). In our results on real data, we
show the performance gain for a standard dimensionality-reduction
classification pipeline on the Pavia Centre, Pavia University and Salinas datasets. It turns
out that synthetic samples consistently increase the OA, AA and kappa coefficient. After a performance jump when adding few
features, the (improved) performance remains relatively stable when adding
further synthetic features. Thus, the choice for the exact number of added
features is not critical.
Quantitatively, the exact performance improvement depends on the details of the
processing chain and on the dataset. The mean improvement in our experiments is
$4.5\%$, with variations between one percent and almost ten percent.
These results are encouraging, as the approach itself is quite straightforward,
and can be smoothly integrated into any classification pipeline.



\bibliographystyle{ieeetr}
\bibliography{ms}


\clearpage\clearpage
\title{GMM-Based Synthetic Samples for Classification of Hyperspectral Images With Limited Training Data\\Supplementary Material}
\maketitle
\setcounter{table}{0}
\renewcommand{\thetable}{\arabic{table}}%
\setcounter{figure}{0}
\renewcommand{\thefigure}{\arabic{figure}}%
\setcounter{section}{0}




\section{Overview}

This document contains the full experimental results to complement the
results in the main text. We report results for all combinations of 
the two considered datasets (Pavia Centre and Salinas), two
dimensionality-reduction methods (PCA and NWFE), two classifiers (random
forests and SVM) with the purpose of either optimizing the classification
parameters or adding synthetic samples. For training the classifiers, we used
limited datasets of either 13, 20, 30, or 40 samples per class.

When considering an unoptimized random forest, default parameters from the
literature are used, i.e., $H=100$ trees with a tree depth of the
square root of the feature dimension, 
$D=\sqrt{d}$.

Furthermore, random forests are oftentimes used in the literature with a
default set of variables rather than optimized parameters. We
conducted each of the classification experiments using two versions of the
classifiers: optimized and unoptimized. In our work, we denote a classifier as being ``unoptimized'' if its parameters are taken from reported values instead of being the results of a training protocol. In contrast, optimized classifiers result from a parameter search. 
In this document, we tabulate the full results to all of our experiments.

\section{Simulated Data}
\begin{figure}[tb]
\centering
\begin{subfigure}[b]{0.23\textwidth}
\centering
\includegraphics[width=\textwidth]{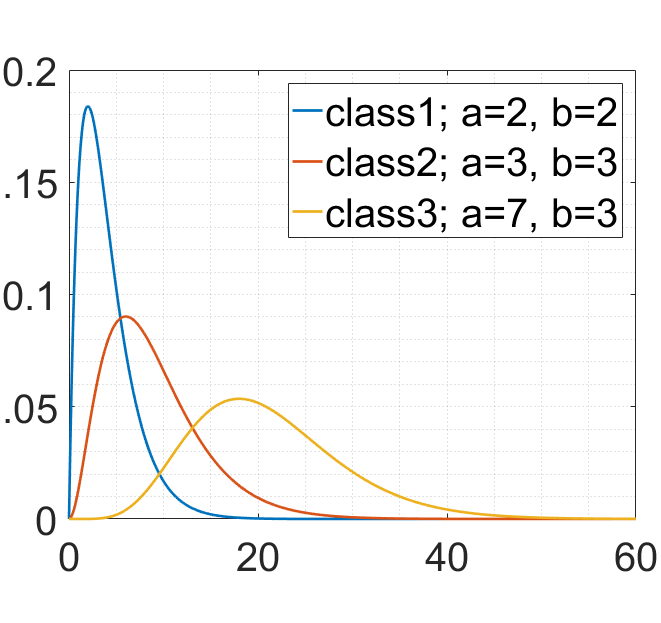}
\caption[]%
{{\small Gamma distributions}}    
\label{gammadists}
\end{subfigure}
\hfill
\begin{subfigure}[b]{0.23\textwidth}   
\centering 
\includegraphics[width=\textwidth]{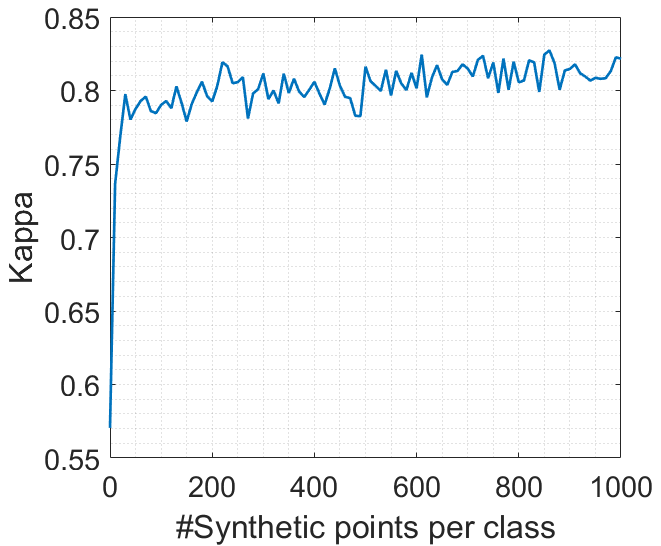}
\caption[]%
{{\small Kappa}}  
\label{gammaKappa}
\end{subfigure}
\caption[  ]
{\small Simulation example on model accuracy versus classifier support: (a) three different gamma distributions serving as the simulated data's class distributions, (b) classification performance (kappa value) vs. number of added synthetic samples drawn from the fitted GMM.} 
\label{gammaress}
\end{figure}

Since we are operating on limited training data, we consider a GMM as a reasonable trade-off between the model complexity and the expressiveness of the available samples.
This is illustrated with a small simulation experiment. We generate three Gamma
distributions with varying shape parameter $a$ and scale parameter $b$. These
distributions are shown in Fig.~\ref{gammadists}. 1000, 2000 and 3000 samples
are drawn from these distributions to simulate a gamma distributed dataset.
From each dataset, 13 samples per class are randomly selected as the training
set, and a GMM is fitted to each of these classes (with the same
parametrization as for the real-world experiments stated further below).
Then, we sample $0 \le n_s \le 1000$ additional training samples from the
estimated GMMs and add these samples to the original training set. A random
forest classifier is trained on these $a + 13$ samples. The parameters for random forest were chosen the same as what is popularly used in the hyperspectral remote sensing image analysis community, i.e. 100 trees ($H$) and square root of number of features as the maximum leaves depth ($D$), as in \cite{dalla2010extended}, \cite{aptoula2016deep}, \cite{guo2011relevance}. Fig.~\ref{gammaress}
shows the Cohen's kappa of the
classification result in dependency of the number of added samples $n_s$. It can
be seen that the addition of very few synthetic samples already boosts
classification performance. These performance gains remain roughly stable for
up to $1000$ additional samples.  The chosen GMM model is not able to
accurately represent the Gamma distributions, which contributes to the fact
that performance never reaches the optimum. However, it is sufficiently
accurate to considerably improve the performance over the baseline with $500$
added samples.

\section{Random Forest + Synthetic Samples}

Tables~\ref{pavia_pca_rf_untuned_numeric_suppl} and \ref{salinas_pca_rf_untuned_numeric_suppl} contain the classification results using the unoptimized random forest classifier. In Tab.~\ref{pavia_pca_rf_untuned_numeric_suppl}, features are computed over Pavia Centre dataset and reduced via PCA and NWFE. 

\begin{table}[tbh]
	\centering
	\caption{Unoptimized random forest-based classification performance on Pavia Centre
	dataset with and without adding synthetic features.}
	\begin{tabular}{|l|c@{ }|c@{ }c@{ }c|}
		\hline
		Algorithm & $|S|$ & AA\% ($\pm$SD) & OA\% ($\pm$SD) & Kappa ($\pm$SD) \\
		\hline
		\hline
		\rowcolor[rgb]{ .706,  .702,  .702} \multicolumn{5}{|c|}{Pavia Centre} \\
		\hline
		\rowcolor[rgb]{ .906,  .902,  .902}\multicolumn{5}{|c|}{13 pix/class} \\
		EMAP & - & 77.87 ($\pm$2.97) & 90.01 ($\pm$3.78) & 0.8600 ($\pm$0.0495)  \\
		\hline
		EMAP-PCA & - &  73.51 ($\pm$3.00) & 86.38 ($\pm$3.61) & 0.8089 ($\pm$0.0493)  \\
		EMAP-PCA & 500 & \textbf{84.59} ($\pm$1.58) & \textbf{93.67} ($\pm$0.75) & \textbf{0.9107} ($\pm$0.0104)  \\
		\hline
		EMAP-NWFE & - & 80.06 ($\pm$3.56) & 91.37 ($\pm$2.67) & 0.8787 ($\pm$0.0365)\\
		EMAP-NWFE & 500 & \textbf{89.57} ($\pm$1.15)  & \textbf{95.91} ($\pm$0.49)  & \textbf{0.9423} ($\pm$0.0069) \\
		\hline
		\rowcolor[rgb]{ .906,  .902,  .902}  \multicolumn{5}{|c|}{20 pix/class} \\ 
		EMAP & - & 81.80 ($\pm$2.07) & 92.73 ($\pm$1.23) & 0.8974 ($\pm$0.0171)  \\
		\hline
		EMAP-PCA & -   & 79.07 ($\pm$1.69) & 90.89 ($\pm$1.22) & 0.8717 ($\pm$0.0169)  \\
		EMAP-PCA & 500 & \textbf{86.37} ($\pm$1.05) & \textbf{94.16} ($\pm$0.47) & \textbf{0.9178} ($\pm$0.0065)  \\
		\hline
		EMAP-NWFE & - & 83.32 ($\pm$2.24) & 93.28 ($\pm$1.30) & 0.9053 ($\pm$0.0181)\\
		EMAP-NWFE & 500 & \textbf{90.68} ($\pm$1.05) & \textbf{96.17} ($\pm$0.50) & \textbf{0.9460} ($\pm$0.0070)\\
		\hline
		\rowcolor[rgb]{ .906,  .902,  .902}  \multicolumn{5}{|c|}{30 pix/class} \\ 
		EMAP & - & 85.57 ($\pm$1.29) & 93.95 ($\pm$0.59) & 0.9148 ($\pm$0.0082)  \\
		\hline
		EMAP-PCA & - & 82.19 ($\pm$1.29) & 92.97 ($\pm$0.75) & 0.9008 ($\pm$0.0105)  \\
		EMAP-PCA & 500 & \textbf{87.91} ($\pm$1.09) & \textbf{94.75} ($\pm$0.48) & \textbf{0.9260} ($\pm$0.0067)  \\
		\hline
		EMAP-NWFE & - & 86.69 ($\pm$1.12)  & 94.89 ($\pm$0.74) & 0.9279 ($\pm$0.0103)\\
		EMAP-NWFE & 500 & \textbf{91.70} ($\pm$0.90) & \textbf{96.53} ($\pm$0.56) & \textbf{0.9510} ($\pm$0.0078)\\
		\hline
		\rowcolor[rgb]{ .906,  .902,  .902}  \multicolumn{5}{|c|}{40 pix/class} \\  
		EMAP & - & 86.80 ($\pm$1.47) & 94.30 ($\pm$0.61) & 0.9197 ($\pm$0.0086)  \\
		\hline
		EMAP-PCA & - & 83.98 ($\pm$1.18) & 93.49 ($\pm$0.69) & 0.9082 ($\pm$0.0096)  \\
		EMAP-PCA & 500 & \textbf{88.74} ($\pm$0.96) & \textbf{95.09} ($\pm$0.54) & \textbf{0.9307} ($\pm$0.0076)  \\
		\hline
		EMAP-NWFE & - & 87.41 ($\pm$1.41)  & 95.18 ($\pm$0.61)  & 0.9318 ($\pm$0.0085) \\
		EMAP-NWFE & 500 & \textbf{92.39} ($\pm$0.75) & \textbf{96.66} ($\pm$0.46) & \textbf{0.9528} ($\pm$0.0063)\\
		\hline
	\end{tabular}%
	\label{pavia_pca_rf_untuned_numeric_suppl}%
\end{table}%

Analogously, Tab.~\ref{salinas_pca_rf_untuned_numeric_suppl} presents unoptimized random forest performance for features from EMAP, EMAP-PCA and EMAP-NWFE, with and without synthetic samples, computed over Salinas dataset. $|S|$ represents the number of added synthetic samples. 

\begin{table}[t!]
	\centering
	\caption{Unoptimized random forest performance on Salinas dataset with and without adding synthetic features.}
	\begin{tabular}{|l|c@{ }|c@{ }c@{ }c|}
		\hline
		Algorithm & $|S|$ & AA\% ($\pm$SD) & OA\% ($\pm$SD) & Kappa ($\pm$SD) \\
		\hline
		\hline
		\rowcolor[rgb]{ .706,  .702,  .702} \multicolumn{5}{|c|}{Salinas} \\
		\hline
		\rowcolor[rgb]{ .906,  .902,  .902} \multicolumn{5}{|c|}{13 pix/class} \\ 
		EMAP    & - & 83.84 ($\pm$2.06) & 76.30 ($\pm$2.74) & 0.7380 ($\pm$0.0292)  \\
		\hline
		EMAP-PCA & - & 82.50 ($\pm$2.06) & 74.96 ($\pm$3.63) & 0.7230 ($\pm$0.0378)  \\
		EMAP-PCA & 500 & \textbf{90.96} ($\pm$0.88) & \textbf{83.89} ($\pm$1.72) & \textbf{0.8215} ($\pm$0.0188)  \\
		\hline
		EMAP-NWFE & - & 88.68 ($\pm$1.20)  & 80.42 ($\pm$2.34)  & 0.7838 ($\pm$0.0247)  \\
		EMAP-NWFE & 500 & \textbf{93.17} ($\pm$0.68) & \textbf{87.09} ($\pm$1.26) & \textbf{0.8566} ($\pm$0.0138)  \\
		\hline
		\rowcolor[rgb]{ .906,  .902,  .902} \multicolumn{5}{|c|}{20 pix/class} \\ 
		EMAP    & - & 86.81 ($\pm$1.63) & 79.74 ($\pm$2.56) & 0.7756 ($\pm$0.0269)  \\
		\hline
		EMAP-PCA & - & 86.59 ($\pm$1.06) & 78.70 ($\pm$2.33) & 0.7643 ($\pm$0.0249)  \\
		EMAP-PCA & 500 & \textbf{91.76} ($\pm$0.97) & \textbf{85.38} ($\pm$1.40) & \textbf{0.8376} ($\pm$0.0154)  \\
		\hline
		EMAP-NWFE & - & 90.56 ($\pm$1.26)  & 82.26 ($\pm$2.62)  & 0.8038 ($\pm$0.0280)  \\
		EMAP-NWFE & 500 & \textbf{94.00} ($\pm$0.39) & \textbf{88.28} ($\pm$1.01) & \textbf{0.8697} ($\pm$0.0111)  \\
		\hline
		\rowcolor[rgb]{ .906,  .902,  .902} \multicolumn{5}{|c|}{30 pix/class} \\ 
		EMAP    & - & 89.01 ($\pm$1.10) & 81.80 ($\pm$2.30) & 0.7985 ($\pm$0.0248)  \\
		\hline
		EMAP-PCA & - & 88.85 ($\pm$0.91) & 80.96 ($\pm$2.15) & 0.7895 ($\pm$0.0229)  \\
		EMAP-PCA & 500 & \textbf{92.63} ($\pm$0.47) & \textbf{86.27} ($\pm$0.94) & \textbf{0.8476} ($\pm$0.0103)  \\
		\hline
		EMAP-NWFE & - & 92.25 ($\pm$0.82)  & 84.76 ($\pm$2.38)  & 0.8314 ($\pm$0.0256)  \\
		EMAP-NWFE & 500 & \textbf{94.35} ($\pm$0.43) & \textbf{88.74} ($\pm$1.24) & \textbf{0.8748} ($\pm$0.0138)  \\
		\hline
		\rowcolor[rgb]{ .906,  .902,  .902} \multicolumn{5}{|c|}{40 pix/class} \\ 
		EMAP    & - & 90.75 ($\pm$0.86) & 84.52 ($\pm$1.76) & 0.8285 ($\pm$0.0192)  \\
		\hline
		EMAP-PCA & - & 89.80 ($\pm$1.21) & 81.73 ($\pm$2.51) & 0.7981 ($\pm$0.0273)  \\
		EMAP-PCA & 500 & \textbf{93.08} ($\pm$0.40) & \textbf{86.59} ($\pm$0.85) & \textbf{0.8512} ($\pm$0.0093)  \\
		\hline
		EMAP-NWFE & - & 93.29 ($\pm$0.41)  & 86.09 ($\pm$1.86)  & 0.8462 ($\pm$0.0200)  \\
		EMAP-NWFE & 500 & \textbf{94.52} ($\pm$0.30) & \textbf{89.18} ($\pm$0.81) & \textbf{0.8798} ($\pm$0.0089)  \\
		\hline
	\end{tabular}%
	\label{salinas_pca_rf_untuned_numeric_suppl}%
\end{table}%

As it was mentioned in the main paper, to further evaluate our idea and compare with other works, we conducted our experiments on the commonly used Pavia University dataset. Tab.~\ref{paviau_pca_rf_untuned_numeric_suppl} presents unoptimized random forest performance for features from EMAP, EMAP-PCA and EMAP-NWFE, with and without synthetic samples, computed over Pavia University dataset. $|S|$ represents the number of added synthetic samples. 

\begin{table}[htbp]
	\centering
	\caption{Unoptimized random forest performance on Pavia University dataset with and without adding synthetic features.}
	\begin{tabular}{|l|c@{ }|c@{ }c@{ }c|}
		\hline
		Algorithm & $|S|$ & AA\% ($\pm$SD) & OA\% ($\pm$SD) & Kappa ($\pm$SD) \\
		\hline
		\hline
		\rowcolor[rgb]{.706,  .702,  .702} \multicolumn{5}{|c|}{Pavia University} \\
		\hline
		\rowcolor[rgb]{ .906,  .902,  .902} \multicolumn{5}{|c|}{13 pix/class} \\
		EMAP  & -     & 70.50 ($\pm$2.88) &  54.81 ($\pm$7.72) &  0.4625 ($\pm$0.0734) \\
		\hline
		EMAP-PCA & -     & 73.77 ($\pm$3.56) &  65.59 ($\pm$8.52) &  0.5726 ($\pm$0.0879) \\
		EMAP-PCA & 500   & \textbf{84.78} ($\pm$1.49) &  \textbf{79.52} ($\pm$2.98) &  \textbf{0.7376} ($\pm$0.0344) \\
		\hline
		EMAP-NWFE & -     & 68.33 ($\pm$2.97) &  62.90 ($\pm$6.89) &  0.5369 ($\pm$0.0671) \\
		EMAP-NWFE & 500   & \textbf{82.87} ($\pm$1.09) &  \textbf{76.13} ($\pm$3.12) &  \textbf{0.6984} ($\pm$0.0340) \\
		\hline
		\rowcolor[rgb]{ .906,  .902,  .902} \multicolumn{5}{|c|}{20 pix/class} \\
		EMAP  & -     & 75.93 ($\pm$2.00) &  65.12 ($\pm$4.74) &  0.5691 ($\pm$0.0467) \\
		\hline
		EMAP-PCA & -     & 78.73 ($\pm$2.28) &  74.87 ($\pm$5.94) &  0.6768 ($\pm$0.0621) \\
		EMAP-PCA & 500   & \textbf{86.75} ($\pm$1.16) &  \textbf{82.24} ($\pm$3.13) &  \textbf{0.7717} ($\pm$0.0364) \\
		\hline
		EMAP-NWFE & -     & 76.87 ($\pm$2.13) &  72.13 ($\pm$7.01) &  0.6466 ($\pm$0.0715) \\
		EMAP-NWFE & 500   & \textbf{83.96} ($\pm$1.20) &  \textbf{77.84} ($\pm$2.57) &  \textbf{0.7183} ($\pm$0.0291) \\
		\hline
		\rowcolor[rgb]{ .906,  .902,  .902} \multicolumn{5}{|c|}{30 pix/class} \\
		EMAP  & -     & 79.63 ($\pm$1.56) &  70.94 ($\pm$3.55) &  0.6353 ($\pm$0.0375) \\
		\hline
		EMAP-PCA & -     & 83.04 ($\pm$1.46) &  77.76 ($\pm$4.75) &  0.7156 ($\pm$0.0495) \\
		EMAP-PCA & 500   & \textbf{87.55} ($\pm$0.87) &  \textbf{83.05} ($\pm$2.29) &  \textbf{0.7818} ($\pm$0.0266) \\
		\hline
		EMAP-NWFE & -     & 80.08 ($\pm$1.68) &  75.16 ($\pm$4.95) &  0.6819 ($\pm$0.0528) \\
		EMAP-NWFE & 500   & \textbf{85.61} ($\pm$0.83) &  \textbf{80.44} ($\pm$2.56) &  \textbf{0.7502} ($\pm$0.0296) \\
		\hline
		\rowcolor[rgb]{ .906,  .902,  .902} \multicolumn{5}{|c|}{40 pix/class} \\
		EMAP  & -     & 81.72 ($\pm$1.52) &  71.16 ($\pm$3.61) &  0.6422 ($\pm$0.0379) \\
		\hline
		EMAP-PCA & -     & 84.85 ($\pm$1.10) &  79.71 ($\pm$3.42) &  0.7392 ($\pm$0.0381) \\
		EMAP-PCA & 500   & \textbf{88.36} ($\pm$1.03) &  \textbf{83.72} ($\pm$2.27) &  \textbf{0.7907} ($\pm$0.0266) \\
		\hline
		EMAP-NWFE & -     & 81.71 ($\pm$1.35) &  76.47 ($\pm$4.36) &  0.6996 ($\pm$0.0473) \\
		EMAP-NWFE & 500   & \textbf{86.07} ($\pm$0.93) &  \textbf{80.86} ($\pm$2.85) &  \textbf{0.7557} ($\pm$0.0329) \\
		\hline
	\end{tabular}%
	\label{paviau_pca_rf_untuned_numeric_suppl}%
\end{table}%


\section{SVM + Synthetic Samples}

Tab.~\ref{pavia_pca_svm_untuned_numeric_suppl} shows the classification
results of EMAP, EMAP-PCA, EMAP-NWFE and variants thereof with added synthetic samples, computed over
Pavia Centre dataset, using an unoptimized SVM classifier. Similarly, Tab.~\ref{salinas_pca_svm_untuned_numeric_suppl} exhibits the unoptimized SVM
 results of the same features,
 but computed over Salinas datasets.

\begin{table}[tbh]
	\centering
	\caption{Unoptimized SVM performance on Pavia Centre with and without adding synthetic features. Parameters of SVM are arbitrarily chosen to be $C=1$ and $\gamma = 5$ for both EMAP-PCA and EMAP-NWFE datasets.}
	\begin{tabular}{|l|c@{ }|c@{ }c@{ }c|}
		\hline
		Algorithm & $|S|$ & AA\% ($\pm$SD) & OA\% ($\pm$SD) & Kappa ($\pm$SD) \\
		\hline
		\hline
		\rowcolor[rgb]{ .706,  .702,  .702} \multicolumn{5}{|c|}{Pavia Centre} \\
		\hline
		\rowcolor[rgb]{ .906,  .902,  .902} \multicolumn{5}{|c|}{13 pix/class} \\
		EMAP & - & 89.77 ($\pm$1.75) & 95.37 ($\pm$0.89) & 0.9348 ($\pm$0.0124) \\
		\hline
		EMAP-PCA & - & 75.11 ($\pm$2.21) & 90.41 ($\pm$1.28) & 0.8627 ($\pm$0.0188) \\
		EMAP-PCA & 500 & \textbf{87.01} ($\pm$1.08) & \textbf{94.68} ($\pm$0.48) & \textbf{0.9249} ($\pm$0.0067) \\
		\hline
		EMAP-NWFE & - & 77.27 ($\pm$1.69) & 91.96 ($\pm$1.09) & 0.8851 ($\pm$0.0157) \\
		EMAP-NWFE & 500 & \textbf{90.01} ($\pm$0.87) & \textbf{95.32} ($\pm$0.35) & \textbf{0.9341} ($\pm$0.0049) \\
		\hline
		\rowcolor[rgb]{ .906,  .902,  .902} \multicolumn{5}{|c|}{20 pix/class} \\
		EMAP & - & 90.92 ($\pm$1.23) & 96.27 ($\pm$0.52) & 0.9473 ($\pm$0.0073) \\
		\hline
		EMAP-PCA & - & 78.45 ($\pm$1.42) & 91.87 ($\pm$0.72) & 0.8844 ($\pm$0.0105) \\
		EMAP-PCA & 500 & \textbf{87.51} ($\pm$1.42) & \textbf{94.96} ($\pm$0.50) & \textbf{0.9289} ($\pm$0.0071) \\
		\hline
		EMAP-NWFE & - & 78.91 ($\pm$1.49) & 92.37 ($\pm$0.71) & 0.8915 ($\pm$0.0102) \\
		EMAP-NWFE & 500 & \textbf{91.14} ($\pm$0.64) & \textbf{95.74} ($\pm$0.29) & \textbf{0.9400} ($\pm$0.0040) \\
		\hline
		\rowcolor[rgb]{ .906,  .902,  .902} \multicolumn{5}{|c|}{30 pix/class} \\
		EMAP & - & 93.31 ($\pm$0.59) & 97.01 ($\pm$0.37) & 0.9578 ($\pm$0.0052) \\
		\hline
		EMAP-PCA & - & 80.79 ($\pm$1.71) & 92.80 ($\pm$0.83) & 0.8977 ($\pm$0.0119) \\
		EMAP-PCA & 500 & \textbf{89.05} ($\pm$0.39) & \textbf{95.63} ($\pm$0.36) & \textbf{0.9382} ($\pm$0.0049) \\
		\hline
		EMAP-NWFE & - & 82.63 ($\pm$1.23) & 94.40 ($\pm$0.52) & 0.9206 ($\pm$0.0073) \\
		EMAP-NWFE & 500 & \textbf{91.15} ($\pm$0.76) & \textbf{96.41} ($\pm$0.09) & \textbf{0.9493} ($\pm$0.0013) \\
		\hline
		\rowcolor[rgb]{ .906,  .902,  .902} \multicolumn{5}{|c|}{40 pix/class} \\
		EMAP & - & 93.74 ($\pm$0.61) & 97.05 ($\pm$0.43) & 0.9584 ($\pm$0.0060) \\
		\hline
		EMAP-PCA & - & 83.52 ($\pm$1.38) & 93.95 ($\pm$0.55) & 0.9142 ($\pm$0.0078) \\
		EMAP-PCA & 500 & \textbf{88.83} ($\pm$0.83) & \textbf{95.39} ($\pm$0.25) & \textbf{0.9348} ($\pm$0.0035) \\
		\hline
		EMAP-NWFE & - & 83.52 ($\pm$0.98) & 93.99 ($\pm$0.42) & 0.9150 ($\pm$0.0060) \\
		EMAP-NWFE & 500 & \textbf{92.13} ($\pm$0.31) & \textbf{96.42} ($\pm$0.31) & \textbf{0.9495} ($\pm$0.0043) \\
		\hline
	    \end{tabular}%
	\label{pavia_pca_svm_untuned_numeric_suppl}%
\end{table}%


\begin{table}[tbh]
	\centering
	\caption{Unoptimized SVM performance on Salinas with and without adding synthetic features. Parameters of SVM are arbitrarily chosen to be $C=1$ and $\gamma = 1$ for EMAP-PCA dataset and $C=1$ and $\gamma = 5$ for EMAP-NWFE dataset.}
	\begin{tabular}{|l|c@{ }|c@{ }c@{ }c|}
		\hline
		Algorithm & $|S|$ & AA\% ($\pm$SD) & OA\% ($\pm$SD) & Kappa ($\pm$SD) \\
		\hline
		\hline
		\rowcolor[rgb]{ .706,  .702,  .702} \multicolumn{5}{|c|}{Salinas} \\
		\hline
		\rowcolor[rgb]{ .906,  .902,  .902} \multicolumn{5}{|c|}{13 pix/class} \\
		EMAP  & - & 89.47 ($\pm$1.13) & 80.82 ($\pm$1.33) & 0.7886 ($\pm$0.0146) \\
		\hline
		EMAP-PCA & - & 88.45 ($\pm$0.50) & 79.60 ($\pm$1.95) & 0.7749 ($\pm$0.0204) \\
		EMAP-PCA & 500 & \textbf{92.69} ($\pm$0.74) & \textbf{85.74} ($\pm$1.94) & \textbf{0.8420} ($\pm$0.0212) \\
		\hline
		EMAP-NWFE & - & 84.23 ($\pm$1.32) & 75.21 ($\pm$3.23) & 0.7266 ($\pm$0.0332) \\
		EMAP-NWFE & 500 & \textbf{93.21} ($\pm$0.69) & \textbf{86.34} ($\pm$1.83) & \textbf{0.8485} ($\pm$0.0202) \\
		\hline
		\rowcolor[rgb]{ .906,  .902,  .902} \multicolumn{5}{|c|}{20 pix/class} \\
		EMAP  & - & 91.17 ($\pm$0.66) & 83.47 ($\pm$1.12) & 0.8174 ($\pm$0.0123) \\
		\hline
		EMAP-PCA & - & 89.87 ($\pm$0.83) & 83.00 ($\pm$1.74) & 0.8119 ($\pm$0.0187) \\
		EMAP-PCA & 500 & \textbf{92.75} ($\pm$0.94) & \textbf{85.90} ($\pm$1.63) & \textbf{0.8438} ($\pm$0.0179) \\
		\hline
		EMAP-NWFE & - & 84.15 ($\pm$0.64) & 75.91 ($\pm$2.12) & 0.7346 ($\pm$0.0223) \\
		EMAP-NWFE & 500 & \textbf{93.96} ($\pm$0.29) & \textbf{87.69} ($\pm$1.28) & \textbf{0.8635} ($\pm$0.0140) \\
		\hline
		\rowcolor[rgb]{ .906,  .902,  .902} \multicolumn{5}{|c|}{30 pix/class} \\
		EMAP  & - & 92.31 ($\pm$0.54) & 84.80 ($\pm$0.92) & 0.8320 ($\pm$0.0100) \\
		\hline
		EMAP-PCA & - & 90.64 ($\pm$0.91) & 82.06 ($\pm$2.23) & 0.8021 ($\pm$0.0239) \\
		EMAP-PCA & 500 & \textbf{93.76} ($\pm$0.44) & \textbf{87.63} ($\pm$0.70) & \textbf{0.8628} ($\pm$0.0078) \\
		\hline
		EMAP-NWFE & - & 84.80 ($\pm$0.52) & 75.20 ($\pm$2.31) & 0.7270 ($\pm$0.0237) \\
		EMAP-NWFE & 500 & \textbf{93.94} ($\pm$0.32) & \textbf{88.57} ($\pm$0.35) & \textbf{0.8730} ($\pm$0.0039) \\
		\hline
		\rowcolor[rgb]{ .906,  .902,  .902} \multicolumn{5}{|c|}{40 pix/class} \\
		EMAP  & - & 92.88 ($\pm$0.45) & 85.79 ($\pm$1.23) & 0.8429 ($\pm$0.0134) \\
		\hline
		EMAP-PCA & - & 91.40 ($\pm$0.93) & 83.60 ($\pm$2.47) & 0.8190 ($\pm$0.0266) \\
		EMAP-PCA & 500 & \textbf{93.84} ($\pm$0.50) & \textbf{87.67} ($\pm$1.25) & \textbf{0.8632} ($\pm$0.0137) \\
		\hline
		EMAP-NWFE & - & 87.08 ($\pm$2.29) & 78.00 ($\pm$4.44) & 0.7580 ($\pm$0.0476) \\
		EMAP-NWFE & 500 & \textbf{94.13} ($\pm$0.33) & \textbf{88.70} ($\pm$0.83) & \textbf{0.8745} ($\pm$0.0092) \\
		\hline
	\end{tabular}%
	\label{salinas_pca_svm_untuned_numeric_suppl}%
\end{table}%

\section{Classifier Parameter Selection}\label{subsec:param_selection}
When using, e.g., the support vector machine classifier (SVM), it is widely
known that parameter selection is a critical preparatory step towards obtaining competitive
results. This is the reason why, for example, SVM parameter selection is 
hardwired into the popular SVM implementation \texttt{libSVM}.
However, other classification frameworks do not
necessarily include a parameter selection submodule. 
One notable example is classification with a random forest.
Several works~\cite{dalla2010morphological,guo2011relevance,aptoula2015hyperspectral,aptoula2014impact,aptoula2016vector,davari2015effect} rely on
the default settings of $100$ trees with a tree depth equal to the square root
of the feature dimensionality, $\sqrt{d}$, as originally proposed by
Breiman~\cite{breiman2001random}. However, these parameters have been proposed
based on training on a relatively large dataset.
In the case of classification on severely limited training data, such
default parameters yield suboptimal classification performance.

\begin{figure}[tb]
	\centering
	\includegraphics[width=0.49\linewidth]{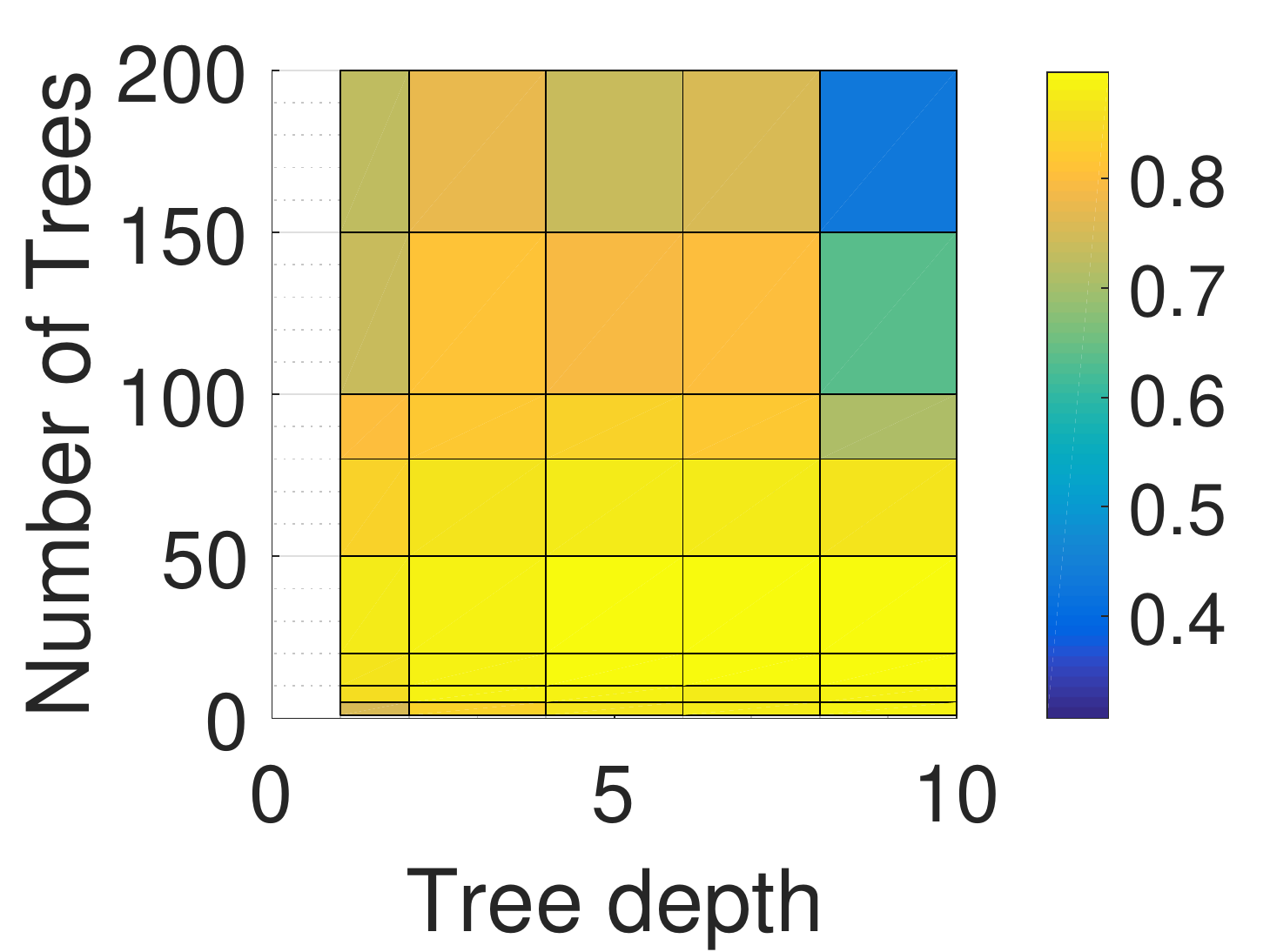}
	\includegraphics[width=0.49\linewidth]{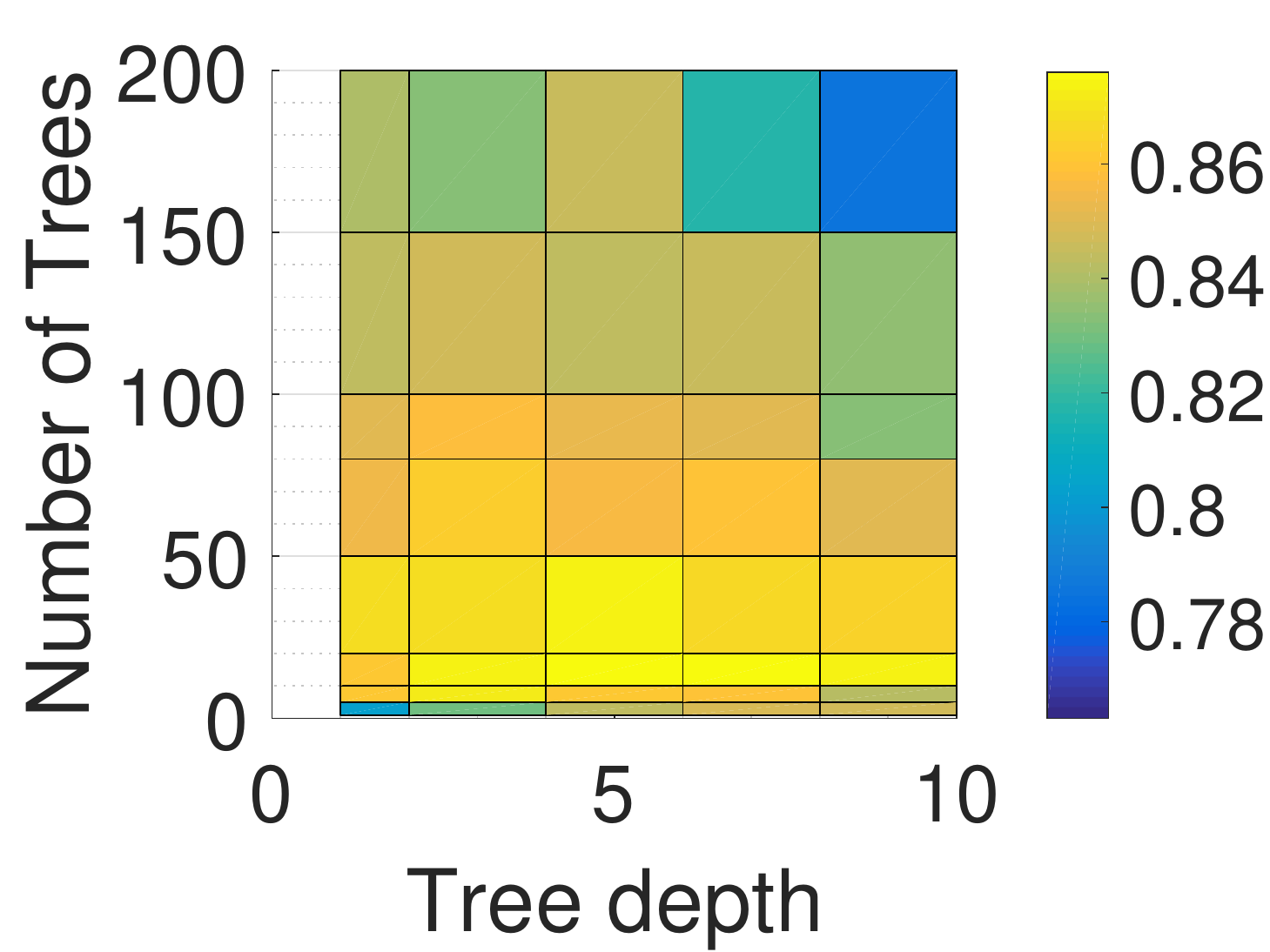}
	\caption{Random forest kappa performance using different number of trees $H$
		and tree depths $D$. Left: Pavia-EMAP-PCA with training set size of $13$ pixels
		per class. Right: Salinas-EMAP-NWFE with training set size of $40$ pixels
		per class.} 
	\label{fig:rf_height_depth_illustration}
\end{figure}

To illustrate how far off the default parameters can be from the optimum
solution, we show two example results in
Fig.~\ref{fig:rf_height_depth_illustration}. A similar study for large training
sets has been done by Rodriguez-Galiano~\emph{et
	al.}\cite{rodriguez2012assessment}. We used limited training sets of size $13$
and $40$, respectively. The features are extracted from Pavia Centre dataset
using PCA-reduced EMAP features, and from Salinas dataset using NWFE-reduced
EMAP features, respectively. 
We color-code the kappa classification performance
for different random forest configurations, i.e., different numbers of trees
and
tree depths.
In both examples, considerably smaller number of trees perform
significantly better.

\subsection{Optimized Parameters for Random Forest}\label{Tuned_RF_results}


Classification performance of optimized random forest classifier on EMAP and EMAP-reduced features computed over Pavia Centre and Salinas dataset are shown in Tab.~\ref{pavia_rf_tuned_numeric_suppl} and Tab.~\ref{salinas_rf_tuned_numeric_suppl}, respectively. The optimized parameters of random forest, $H$ and $D$, are also listed in the tables. 

\begin{table}[tbh]
	\centering
	\caption{Optimized random forest-based classification performance on Pavia Centre dataset without adding synthetic features. $H$ and $D$ represent the tuned parameters of random forest.}
	\begin{tabular}{|l@{ }|c@{ }c@{ }|c@{ }c@{ }c|}
		\hline
		Algorithm & $H$ & $D$ & AA\% ($\pm$SD) & OA\% ($\pm$SD) & Kappa ($\pm$SD) \\
		\hline
		\hline
		\rowcolor[rgb]{ .906,  .902,  .902} \multicolumn{6}{|c|}{13 pix/class} \\
		EMAP  & 5     & 4     & 86.55 ($\pm$1.73) & 93.99 ($\pm$1.11) & 0.9154 ($\pm$0.0151) \\
		EMAP-PCA & 10    & 8     & 84.05 ($\pm$1.44) & 92.67 ($\pm$0.82) & 0.8970 ($\pm$0.0113) \\
		EMAP-NWFE & 10    & 4     & 87.38 ($\pm$1.04) & 94.86 ($\pm$0.47) & 0.9276 ($\pm$0.0066) \\
		\hline
		\rowcolor[rgb]{ .906,  .902,  .902} \multicolumn{6}{|c|}{20 pix/class} \\
		EMAP  & 10    & 10    & 89.59 ($\pm$1.24) & 94.94 ($\pm$0.76) & 0.9289 ($\pm$0.0106) \\
		EMAP-PCA & 20    & 10    & 86.21 ($\pm$1.24) & 93.80 ($\pm$0.53) & 0.9128 ($\pm$0.0074) \\
		EMAP-NWFE & 20    & 2     & 88.01 ($\pm$0.97) & 95.21 ($\pm$0.56) & 0.9325 ($\pm$0.0079) \\
		\hline
		\rowcolor[rgb]{ .906,  .902,  .902} \multicolumn{6}{|c|}{30 pix/class} \\
		EMAP  & 10    & 6     & 90.84 ($\pm$1.08) & 95.60 ($\pm$0.62) & 0.9381 ($\pm$0.0087) \\
		EMAP-PCA & 10    & 8     & 87.47 ($\pm$0.79) & 94.22 ($\pm$0.38) & 0.9187 ($\pm$0.0053) \\
		EMAP-NWFE & 20    & 4     & 90.16 ($\pm$0.92) & 95.86 ($\pm$0.51) & 0.9416 ($\pm$0.0072) \\
		\hline
		\rowcolor[rgb]{ .906,  .902,  .902} \multicolumn{6}{|c|}{40 pix/class} \\
		EMAP  & 10    & 10    & 92.67 ($\pm$0.74) & 96.03 ($\pm$0.73) & 0.9441 ($\pm$0.0101) \\
		EMAP-PCA & 10    & 6     & 88.97 ($\pm$0.97) & 94.84 ($\pm$0.67) & 0.9274 ($\pm$0.0093) \\
		EMAP-NWFE & 20    & 2     & 90.90 ($\pm$0.96) & 96.17 ($\pm$0.48) & 0.9459 ($\pm$0.0068) \\
		\hline
	\end{tabular}%
	\label{pavia_rf_tuned_numeric_suppl}%
\end{table}%

\begin{table}[tbh]
	\centering
	\caption{Optimized random forest-based classification performance on Salinas dataset without adding synthetic features. $H$ and $D$ represent the tuned parameters of random forest.}
	\begin{tabular}{|l|c@{ }c@{ }|c@{ }c@{ }c|}
		\hline
		Algorithm & $H$ & $D$ & AA\% ($\pm$SD) & OA\% ($\pm$SD) & Kappa ($\pm$SD) \\
		\hline
		\hline
		\rowcolor[rgb]{ .906,  .902,  .902} \multicolumn{6}{|c|}{13 pix/class} \\
		EMAP  & 5     & 10    & 91.54 ($\pm$1.06) & 86.45 ($\pm$2.25) & 0.8496 ($\pm$0.0250) \\
		EMAP-PCA & 10    & 6     & 89.29 ($\pm$1.10) & 82.66 ($\pm$1.39) & 0.8077 ($\pm$0.0153) \\
		EMAP-NWFE & 10    & 4     & 91.85 ($\pm$0.89) & 85.21 ($\pm$1.18) & 0.8357 ($\pm$0.0132) \\
		\hline
		\rowcolor[rgb]{ .906,  .902,  .902} \multicolumn{6}{|c|}{20 pix/class} \\
		EMAP  & 5     & 10    & 93.10 ($\pm$0.93) & 88.13 ($\pm$1.86) & 0.8683 ($\pm$0.0206) \\
		EMAP-PCA & 10    & 10    & 90.68 ($\pm$0.73) & 84.14 ($\pm$1.21) & 0.8241 ($\pm$0.0133) \\
		EMAP-NWFE & 20    & 6     & 92.75 ($\pm$0.54) & 86.58 ($\pm$0.94) & 0.8511 ($\pm$0.0103) \\
		\hline
		\rowcolor[rgb]{ .906,  .902,  .902} \multicolumn{6}{|c|}{30 pix/class} \\
		EMAP  & 5     & 10    & 94.33 ($\pm$0.93) & 90.32 ($\pm$1.40) & 0.8926 ($\pm$0.0155) \\
		EMAP-PCA & 10    & 10    & 91.67 ($\pm$0.73) & 85.58 ($\pm$0.68) & 0.8399 ($\pm$0.0076) \\
		EMAP-NWFE & 10    & 4     & 93.83 ($\pm$0.43) & 88.11 ($\pm$0.92) & 0.8679 ($\pm$0.0101) \\
		\hline
		\rowcolor[rgb]{ .906,  .902,  .902} \multicolumn{6}{|c|}{40 pix/class} \\
		EMAP  & 5     & 10    & 95.13 ($\pm$0.71) & 91.55 ($\pm$1.08) & 0.9062 ($\pm$0.0119) \\
		EMAP-PCA & 10    & 4     & 92.43 ($\pm$0.33) & 86.38 ($\pm$0.52) & 0.8489 ($\pm$0.0058) \\
		EMAP-NWFE & 10    & 6     & 94.20 ($\pm$0.30) & 88.83 ($\pm$0.88) & 0.8760 ($\pm$0.0097) \\
		\hline
	\end{tabular}%
	\label{salinas_rf_tuned_numeric_suppl}%
\end{table}%

\subsection{Optimized Parameters for SVM}
Analogously to Sec.~\ref{Tuned_RF_results}, Tab.~\ref{pavia_svm_tuned_numeric_suppl} and Tab.~\ref{salinas_svm_tuned_numeric_suppl} show the classification results using an optimized SVM on Pavia Centre and Salinas datasets, respectively. As SVM classifier is by design a two class classifier, we use a one-versus-all approach to multi-class classification. For classifier tuning, the parameters for each classifier result from a grid-search. Thus, each classifier obtains a unique set of parameters $C$ and $\gamma$. Therefore, there is not a single best set of parameters for the overall classifier, which is why these parameters are not reported here. 

\begin{table}[tbh]
	\centering
	\caption{Tuned SVM-based classification performance of EMAP, EMAP-PCA and EMAP-NWFE computed over Pavia Centre dataset without adding synthetic features.}
	\begin{tabular}{|l|c@{ }c@{ }c|}
		\hline
		\multicolumn{1}{|l|}{Algorithm} & AA\% ($\pm$SD) & OA\% ($\pm$SD) & Kappa ($\pm$SD) \\
		\hline
		\hline
		\rowcolor[rgb]{ .906,  .902,  .902} \multicolumn{4}{|c|}{13 pix/class} \\
		EMAP  & 87.20 ($\pm$1.53) & 94.24 ($\pm$1.07) & 0.9189 ($\pm$0.0148) \\
		EMAP-PCA & 86.21 ($\pm$2.83) & 93.85 ($\pm$1.10) & 0.9134 ($\pm$0.0154) \\
		EMAP-NWFE & 88.42 ($\pm$1.48) & 94.86 ($\pm$0.68) & 0.9276 ($\pm$0.0095) \\
		\hline
		\rowcolor[rgb]{ .906,  .902,  .902} \multicolumn{4}{|c|}{20 pix/class} \\
		EMAP  & 90.77 ($\pm$1.42) & 95.64 ($\pm$0.67) & 0.9385 ($\pm$0.0094) \\
		EMAP-PCA & 88.41 ($\pm$1.35) & 94.82 ($\pm$0.64) & 0.9269 ($\pm$0.0089) \\
		EMAP-NWFE & 90.72 ($\pm$1.51) & 95.60 ($\pm$0.70) & 0.9380 ($\pm$0.0099) \\
		\hline
		\rowcolor[rgb]{ .906,  .902,  .902} \multicolumn{4}{|c|}{30 pix/class} \\
		EMAP  & 92.57 ($\pm$0.85) & 96.16 ($\pm$0.39) & 0.9459 ($\pm$0.0055) \\
		EMAP-PCA & 91.46 ($\pm$0.96) & 95.90 ($\pm$0.57) & 0.9422 ($\pm$0.0080) \\
		EMAP-NWFE & 92.27 ($\pm$0.89) & 96.28 ($\pm$0.47) & 0.9475 ($\pm$0.0065) \\
		\hline
		\rowcolor[rgb]{ .906,  .902,  .902} \multicolumn{4}{|c|}{40 pix/class} \\
		EMAP  & 93.61 ($\pm$0.84) & 96.97 ($\pm$0.52) & 0.9573 ($\pm$0.0073) \\
		EMAP-PCA & 92.55 ($\pm$1.18) & 96.45 ($\pm$0.46) & 0.9500 ($\pm$0.0064) \\
		EMAP-NWFE & 93.46 ($\pm$0.80) & 96.82 ($\pm$0.36) & 0.9551 ($\pm$0.0051) \\
		\hline
	\end{tabular}%
	\label{pavia_svm_tuned_numeric_suppl}%
\end{table}%

\begin{table}[tbh]
	\centering
	\caption{Tuned SVM-based classification performance of EMAP, EMAP-PCA and EMAP-NWFE computed over Salinas dataset without adding synthetic features.}
	\begin{tabular}{|l@{ }|c@{ }c@{ }c@{ }|}
		\hline
		\multicolumn{1}{|l|}{Algorithm} & AA\% ($\pm$SD) & OA\% ($\pm$SD) & Kappa ($\pm$SD) \\
		\hline
		\hline
		\rowcolor[rgb]{ .906,  .902,  .902} \multicolumn{4}{|c|}{13 pix/class} \\
		EMAP  & 93.55 ($\pm$0.60) & 87.41 ($\pm$1.35) & 0.8604 ($\pm$0.0149) \\
		EMAP-PCA & 91.73 ($\pm$1.01) & 84.57 ($\pm$1.96) & 0.8292 ($\pm$0.0213) \\
		EMAP-NWFE & 93.78 ($\pm$0.90) & 87.94 ($\pm$1.88) & 0.8660 ($\pm$0.0208) \\
		\hline
		\rowcolor[rgb]{ .906,  .902,  .902} \multicolumn{4}{|c|}{20 pix/class} \\
		EMAP  & 94.69 ($\pm$0.64) & 89.92 ($\pm$1.44) & 0.8880 ($\pm$0.0160) \\
		EMAP-PCA & 93.02 ($\pm$0.47) & 86.45 ($\pm$1.15) & 0.8497 ($\pm$0.0126) \\
		EMAP-NWFE & 94.86 ($\pm$0.47) & 89.96 ($\pm$1.47) & 0.8885 ($\pm$0.0163) \\
		\hline
		\rowcolor[rgb]{ .906,  .902,  .902} \multicolumn{4}{|c|}{30 pix/class} \\
		EMAP  & 95.28 ($\pm$0.45) & 91.22 ($\pm$1.40) & 0.9025 ($\pm$0.0155) \\
		EMAP-PCA & 93.81 ($\pm$0.48) & 87.57 ($\pm$1.18) & 0.8621 ($\pm$0.0131) \\
		EMAP-NWFE & 95.54 ($\pm$0.49) & 91.91 ($\pm$1.25) & 0.9101 ($\pm$0.0138) \\
		\hline
		\rowcolor[rgb]{ .906,  .902,  .902} \multicolumn{4}{|c|}{40 pix/class} \\
		EMAP  & 95.91 ($\pm$0.48) & 92.20 ($\pm$1.15) & 0.9134 ($\pm$0.0127) \\
		EMAP-PCA & 94.08 ($\pm$0.60) & 87.96 ($\pm$1.58) & 0.8664 ($\pm$0.0173) \\
		EMAP-NWFE & 95.79 ($\pm$0.43) & 92.07 ($\pm$1.07) & 0.9118 ($\pm$0.0120) \\
		\hline
	\end{tabular}%
	\label{salinas_svm_tuned_numeric_suppl}%
\end{table}%

\subsection{Performances for Various Random Forest Parameters}

Fig.~\ref{fig:rf_height_depth_illustration} shows that
the classical parameters which are used in the literature for random forest
classifier, i.e., number of trees $H = 100$, and depth, $D =
\sqrt{d}$, are not the optimal parameter values, particularly for
severely limited training data sizes. This conclusion was drawn based on the
experiments conducted over EMAP-PCA and EMAP-NWFE computed over Pavia Centre
and Salinas datasets. Fig.~\ref{fig:rf_height_depth_illustration} illustrates instances of the numerical analysis
which are presented in Tables~\ref{pavia_rf_tuned_numeric_EMAP_pca_13_suppl},
\ref{pavia_rf_tuned_numeric_EMAP_NWFE_13_suppl},
\ref{salinas_rf_tuned_numeric_EMAP_pca_13_suppl},
\ref{salinas_rf_tuned_numeric_EMAP_NWFE_13_suppl}.

\begin{table}[htbp]
	\centering
	\caption{Kappa-based parameter search results for random forest classifier over EMAP-PCA computed on Pavia Centre dataset.}
	
	\begin{tabular}{|l|cccccc|}
		\hline
		\backslashbox{H}{D} & 1     & 2     & 4     & 6     & 8     & 10 \\
		\hline
		\rowcolor[rgb]{ .906,  .902,  .902} \multicolumn{7}{|c|}{13 pix/class} \\
		
		1     & 0.7525 & 0.8343 & 0.8689 & 0.8766 & 0.8861 & 0.8851 \\
		5     & 0.8594 & 0.8795 & 0.8832 & 0.8784 & 0.8815 & 0.8916 \\
		10    & 0.8661 & 0.8853 & 0.8945 & 0.8926 & \textbf{0.8970} & 0.8967 \\
		20    & 0.8739 & 0.8820 & 0.8927 & 0.8942 & 0.8958 & 0.8922 \\
		50    & 0.8351 & 0.8680 & 0.8723 & 0.8755 & 0.8605 & 0.8546 \\
		80    & 0.8036 & 0.8228 & 0.8353 & 0.8150 & 0.7119 & 0.6854 \\
		100   & 0.7371 & 0.8121 & 0.7921 & 0.7958 & 0.6362 & 0.6398 \\
		150   & 0.7298 & 0.7739 & 0.7318 & 0.7503 & 0.4314 & 0.4827 \\
		200   & 0.7298 & 0.7105 & 0.6883 & 0.7079 & 0.3070 & 0.3065 \\
		\hline
		\rowcolor[rgb]{ .906,  .902,  .902} \multicolumn{7}{|c|}{20 pix/class} \\

		1     & 0.7484 & 0.8587 & 0.8818 & 0.8891 & 0.9002 & 0.9005 \\
		5     & 0.8683 & 0.8951 & 0.9043 & 0.8942 & 0.8971 & 0.8983 \\
		10    & 0.8912 & 0.9047 & 0.9068 & 0.9090 & 0.9040 & 0.9082 \\
		20    & 0.8899 & 0.8999 & 0.9083 & 0.9037 & 0.9026 & \textbf{0.9128} \\
		50    & 0.8854 & 0.8884 & 0.9029 & 0.8974 & 0.8941 & 0.8983 \\
		80    & 0.8543 & 0.8707 & 0.8832 & 0.8803 & 0.8676 & 0.8649 \\
		100   & 0.8475 & 0.8742 & 0.8745 & 0.8574 & 0.8404 & 0.8359 \\
		150   & 0.7959 & 0.8355 & 0.8348 & 0.8515 & 0.7029 & 0.6732 \\
		200   & 0.8049 & 0.8195 & 0.8236 & 0.7921 & 0.5885 & 0.5609 \\
		\hline
		\rowcolor[rgb]{ .906,  .902,  .902} \multicolumn{7}{|c|}{30 pix/class} \\

		1     & 0.7813 & 0.8547 & 0.8957 & 0.8946 & 0.9126 & 0.9142 \\
		5     & 0.8941 & 0.9070 & 0.9094 & 0.9095 & 0.9093 & 0.9095 \\
		10    & 0.8988 & 0.9184 & 0.9139 & 0.9182 & \textbf{0.9187} & 0.9174 \\
		20    & 0.9083 & 0.9130 & 0.9123 & 0.9153 & 0.9131 & 0.9177 \\
		50    & 0.8994 & 0.9067 & 0.9104 & 0.9132 & 0.9074 & 0.9061 \\
		80    & 0.8945 & 0.9043 & 0.9047 & 0.9083 & 0.9010 & 0.9075 \\
		100   & 0.8818 & 0.8941 & 0.8978 & 0.8989 & 0.8968 & 0.8959 \\
		150   & 0.8610 & 0.8884 & 0.8911 & 0.8856 & 0.8438 & 0.8619 \\
		200   & 0.8437 & 0.8572 & 0.8606 & 0.8697 & 0.7828 & 0.7769 \\
		\hline
		\rowcolor[rgb]{ .906,  .902,  .902} \multicolumn{7}{|c|}{40 pix/class} \\

		1     & 0.8319 & 0.8676 & 0.9027 & 0.9111 & 0.9186 & 0.9174 \\
		5     & 0.9004 & 0.9188 & 0.9183 & 0.9176 & 0.9178 & 0.9153 \\
		10    & 0.9101 & 0.9242 & 0.9262 & \textbf{0.9274} & 0.9208 & 0.9247 \\
		20    & 0.9127 & 0.9208 & 0.9215 & 0.9165 & 0.9216 & 0.9249 \\
		50    & 0.9066 & 0.9164 & 0.9157 & 0.9152 & 0.9139 & 0.9114 \\
		80    & 0.8974 & 0.9094 & 0.9132 & 0.9104 & 0.9084 & 0.9084 \\
		100   & 0.8898 & 0.9068 & 0.9099 & 0.9140 & 0.9085 & 0.9026 \\
		150   & 0.8904 & 0.9009 & 0.9060 & 0.9028 & 0.8908 & 0.8879 \\
		200   & 0.8813 & 0.8842 & 0.9018 & 0.8958 & 0.8664 & 0.8563 \\
		\hline
	\end{tabular}%
	\label{pavia_rf_tuned_numeric_EMAP_pca_13_suppl}%
\end{table}%



\begin{table}[htbp]
	\centering
	\caption{Kappa-based parameter search results for random forest classifier over EMAP-NWFE computed on Pavia Centre dataset.}
	\begin{tabular}{|l|cccccc|}
		\hline
		\backslashbox{H}{D} & 1     & 2     & 4     & 6     & 8     & 10 \\
		\hline
		\rowcolor[rgb]{ .906,  .902,  .902} \multicolumn{7}{|c|}{13 pix/class} \\

		1     & 0.7776 & 0.8501 & 0.8576 & 0.8648 & 0.8700 & 0.8607 \\
		5     & 0.8749 & 0.8985 & 0.8779 & 0.8687 & 0.8701 & 0.8811 \\
		10    & 0.8963 & 0.9201 & \textbf{0.9276} & 0.9026 & 0.9147 & 0.9110 \\
		20    & 0.9042 & 0.9213 & 0.9276 & 0.9126 & 0.9183 & 0.9097 \\
		50    & 0.9103 & 0.9057 & 0.9116 & 0.8664 & 0.8702 & 0.8840 \\
		80    & 0.8894 & 0.8848 & 0.8962 & 0.7640 & 0.7882 & 0.7883 \\
		100   & 0.8620 & 0.8924 & 0.8896 & 0.7338 & 0.7662 & 0.7743 \\
		150   & 0.8561 & 0.8819 & 0.8672 & 0.6658 & 0.6528 & 0.6770 \\
		200   & 0.8145 & 0.8298 & 0.8356 & 0.5222 & 0.5605 & 0.5210 \\
		\hline
		\rowcolor[rgb]{ .906,  .902,  .902} \multicolumn{7}{|c|}{20 pix/class} \\

		1     & 0.8265 & 0.8721 & 0.8859 & 0.8800 & 0.8949 & 0.8951 \\
		5     & 0.9025 & 0.9143 & 0.8978 & 0.8912 & 0.9033 & 0.8964 \\
		10    & 0.9175 & 0.9312 & 0.9320 & 0.9199 & 0.9234 & 0.9262 \\
		20    & 0.9139 & \textbf{0.9325} & 0.9322 & 0.9254 & 0.9277 & 0.9221 \\
		50    & 0.9099 & 0.9177 & 0.9259 & 0.9135 & 0.9224 & 0.9129 \\
		80    & 0.9086 & 0.9162 & 0.9228 & 0.8762 & 0.8785 & 0.8826 \\
		100   & 0.9028 & 0.9119 & 0.9174 & 0.8474 & 0.8495 & 0.8471 \\
		150   & 0.8999 & 0.9025 & 0.9029 & 0.7952 & 0.7757 & 0.7639 \\
		200   & 0.8946 & 0.8884 & 0.8914 & 0.7389 & 0.7318 & 0.7202 \\
		\hline
		\rowcolor[rgb]{ .906,  .902,  .902} \multicolumn{7}{|c|}{30 pix/class} \\

		1     & 0.8463 & 0.8877 & 0.9088 & 0.9133 & 0.9076 & 0.9150 \\
		5     & 0.9118 & 0.9290 & 0.9168 & 0.9030 & 0.9107 & 0.9135 \\
		10    & 0.9299 & 0.9379 & 0.9388 & 0.9322 & 0.9334 & 0.9312 \\
		20    & 0.9330 & 0.9377 & \textbf{0.9416} & 0.9337 & 0.9345 & 0.9311 \\
		50    & 0.9276 & 0.9375 & 0.9380 & 0.9288 & 0.9301 & 0.9318 \\
		80    & 0.9205 & 0.9314 & 0.9317 & 0.9195 & 0.9190 & 0.9195 \\
		100   & 0.9205 & 0.9300 & 0.9320 & 0.9124 & 0.9069 & 0.9094 \\
		150   & 0.9203 & 0.9186 & 0.9197 & 0.8725 & 0.8689 & 0.8791 \\
		200   & 0.9154 & 0.9160 & 0.9097 & 0.8019 & 0.7994 & 0.8334 \\
		\hline
		\rowcolor[rgb]{ .906,  .902,  .902} \multicolumn{7}{|c|}{40 pix/class} \\

		1     & 0.8522 & 0.8997 & 0.9095 & 0.9153 & 0.9147 & 0.9181 \\
		5     & 0.9285 & 0.9359 & 0.9288 & 0.9141 & 0.9183 & 0.9243 \\
		10    & 0.9353 & 0.9458 & 0.9375 & 0.9361 & 0.9404 & 0.9356 \\
		20    & 0.9383 & \textbf{0.9459} & 0.9433 & 0.9404 & 0.9393 & 0.9404 \\
		50    & 0.9371 & 0.9413 & 0.9438 & 0.9351 & 0.9364 & 0.9358 \\
		80    & 0.9337 & 0.9356 & 0.9351 & 0.9272 & 0.9316 & 0.9276 \\
		100   & 0.9288 & 0.9379 & 0.9373 & 0.9267 & 0.9296 & 0.9257 \\
		150   & 0.9276 & 0.9319 & 0.9329 & 0.9012 & 0.9101 & 0.9028 \\
		200   & 0.9183 & 0.9268 & 0.9251 & 0.8697 & 0.8786 & 0.8728 \\
		\hline
	\end{tabular}%
	\label{pavia_rf_tuned_numeric_EMAP_NWFE_13_suppl}%
\end{table}%


\begin{table}[htbp]
	\centering
	\caption{Kappa-based parameter search results for random forest classifier over EMAP-PCA computed on Salinas dataset.}
	\begin{tabular}{|l|cccccc|}
		\hline
		\backslashbox{H}{D} & 1     & 2     & 4     & 6     & 8     & 10 \\
		\hline
		\rowcolor[rgb]{ .906,  .902,  .902} \multicolumn{7}{|c|}{13 pix/class} \\

		1     & 0.7350 & 0.7730 & 0.7809 & 0.7840 & 0.7891 & 0.7811 \\
		5     & 0.7963 & 0.7988 & 0.7873 & 0.7851 & 0.7888 & 0.7793 \\
		10    & 0.7919 & 0.7963 & 0.8054 & \textbf{0.8077} & 0.8066 & 0.7956 \\
		20    & 0.7841 & 0.7834 & 0.7883 & 0.7910 & 0.7912 & 0.7935 \\
		50    & 0.7534 & 0.7609 & 0.7596 & 0.7618 & 0.7587 & 0.7512 \\
		80    & 0.7455 & 0.7357 & 0.7024 & 0.7012 & 0.7187 & 0.6991 \\
		100   & 0.7259 & 0.7134 & 0.6567 & 0.6726 & 0.6674 & 0.6526 \\
		150   & 0.6825 & 0.7026 & 0.6043 & 0.5972 & 0.6153 & 0.5930 \\
		200   & 0.6903 & 0.6741 & 0.5520 & 0.5545 & 0.5609 & 0.5497 \\
		\hline
		\rowcolor[rgb]{ .906,  .902,  .902} \multicolumn{7}{|c|}{20 pix/class} \\

		1     & 0.7545 & 0.7876 & 0.8051 & 0.8003 & 0.7994 & 0.7987 \\
		5     & 0.8010 & 0.8135 & 0.7930 & 0.7961 & 0.7993 & 0.8027 \\
		10    & 0.8104 & 0.8184 & 0.8235 & 0.8099 & 0.8164 & \textbf{0.8241} \\
		20    & 0.8013 & 0.8172 & 0.8117 & 0.8030 & 0.8118 & 0.8050 \\
		50    & 0.7849 & 0.7817 & 0.7898 & 0.7951 & 0.7826 & 0.7856 \\
		80    & 0.7699 & 0.7792 & 0.7571 & 0.7621 & 0.7642 & 0.7565 \\
		100   & 0.7508 & 0.7685 & 0.7437 & 0.7472 & 0.7544 & 0.7411 \\
		150   & 0.7352 & 0.7453 & 0.6962 & 0.6921 & 0.6850 & 0.6897 \\
		200   & 0.7251 & 0.7293 & 0.6581 & 0.6392 & 0.6389 & 0.6290 \\
		\hline
		\rowcolor[rgb]{ .906,  .902,  .902} \multicolumn{7}{|c|}{30 pix/class} \\

		1     & 0.7764 & 0.7998 & 0.8174 & 0.8184 & 0.8197 & 0.8155 \\
		5     & 0.8287 & 0.8326 & 0.8179 & 0.8153 & 0.8140 & 0.8137 \\
		10    & 0.8317 & 0.8380 & 0.8380 & 0.8340 & 0.8376 & \textbf{0.8399} \\
		20    & 0.8153 & 0.8287 & 0.8294 & 0.8269 & 0.8323 & 0.8320 \\
		50    & 0.8040 & 0.8055 & 0.8088 & 0.8058 & 0.8123 & 0.8083 \\
		80    & 0.7924 & 0.7881 & 0.7907 & 0.7934 & 0.7900 & 0.7877 \\
		100   & 0.7927 & 0.7861 & 0.7783 & 0.7799 & 0.7791 & 0.7936 \\
		150   & 0.7621 & 0.7704 & 0.7588 & 0.7482 & 0.7570 & 0.7596 \\
		200   & 0.7557 & 0.7523 & 0.7220 & 0.7226 & 0.7123 & 0.7248 \\
		\hline
		\rowcolor[rgb]{ .906,  .902,  .902} \multicolumn{7}{|c|}{40 pix/class} \\

		1     & 0.7904 & 0.8157 & 0.8272 & 0.8216 & 0.8327 & 0.8250 \\
		5     & 0.8350 & 0.8402 & 0.8210 & 0.8282 & 0.8271 & 0.8252 \\
		10    & 0.8396 & 0.8489 & \textbf{0.8489} & 0.8452 & 0.8477 & 0.8437 \\
		20    & 0.8334 & 0.8434 & 0.8425 & 0.8421 & 0.8342 & 0.8422 \\
		50    & 0.8114 & 0.8253 & 0.8103 & 0.8207 & 0.8283 & 0.8262 \\
		80    & 0.8081 & 0.8146 & 0.8059 & 0.8083 & 0.8129 & 0.8114 \\
		100   & 0.8083 & 0.8008 & 0.7962 & 0.8017 & 0.7960 & 0.8022 \\
		150   & 0.7804 & 0.7780 & 0.7876 & 0.7830 & 0.7769 & 0.7899 \\
		200   & 0.7732 & 0.7805 & 0.7639 & 0.7628 & 0.7657 & 0.7703 \\
		\hline
	\end{tabular}%
	\label{salinas_rf_tuned_numeric_EMAP_pca_13_suppl}%
\end{table}%


\begin{table}[htbp]
	\centering
	\caption{Kappa-based parameter search results for random forest classifier over EMAP-NWFE computed on Salinas dataset.}
	\begin{tabular}{|l|cccccc|}
		\hline
		\backslashbox{H}{D} & 1     & 2     & 4     & 6     & 8     & 10 \\
		\hline
		\rowcolor[rgb]{ .906,  .902,  .902} \multicolumn{7}{|c|}{13 pix/class} \\

		1     & 0.7471 & 0.7813 & 0.8001 & 0.7941 & 0.7941 & 0.7956 \\
		5     & 0.8083 & 0.8223 & 0.8221 & 0.8110 & 0.7917 & 0.7933 \\
		10    & 0.8232 & 0.8312 & \textbf{0.8357} & 0.8296 & 0.8258 & 0.8247 \\
		20    & 0.8171 & 0.8171 & 0.8315 & 0.8208 & 0.8220 & 0.8106 \\
		50    & 0.8031 & 0.7893 & 0.7967 & 0.7979 & 0.7547 & 0.7587 \\
		80    & 0.7854 & 0.7755 & 0.7969 & 0.7834 & 0.6499 & 0.6589 \\
		100   & 0.7820 & 0.7817 & 0.7820 & 0.7714 & 0.5493 & 0.5664 \\
		150   & 0.7698 & 0.7677 & 0.7756 & 0.7628 & 0.4025 & 0.4156 \\
		200   & 0.7518 & 0.7458 & 0.7407 & 0.7596 & 0.3003 & 0.3041 \\
		\hline
		\rowcolor[rgb]{ .906,  .902,  .902} \multicolumn{7}{|c|}{20 pix/class} \\

		1     & 0.7434 & 0.8012 & 0.8135 & 0.8203 & 0.8267 & 0.8220 \\
		5     & 0.8337 & 0.8447 & 0.8401 & 0.8294 & 0.8243 & 0.8242 \\
		10    & 0.8349 & 0.8473 & 0.8493 & 0.8497 & 0.8456 & 0.8490 \\
		20    & 0.8370 & 0.8443 & 0.8457 & \textbf{0.8511} & 0.8382 & 0.8345 \\
		50    & 0.8239 & 0.8233 & 0.8238 & 0.8313 & 0.8153 & 0.8165 \\
		80    & 0.8035 & 0.8133 & 0.8060 & 0.8248 & 0.7714 & 0.7744 \\
		100   & 0.7989 & 0.8067 & 0.7946 & 0.7928 & 0.7250 & 0.7318 \\
		150   & 0.7930 & 0.7933 & 0.7995 & 0.8001 & 0.6332 & 0.6135 \\
		200   & 0.7885 & 0.7832 & 0.7755 & 0.7842 & 0.4812 & 0.4911 \\
		\hline
		\rowcolor[rgb]{ .906,  .902,  .902} \multicolumn{7}{|c|}{30 pix/class} \\

		1     & 0.7860 & 0.8146 & 0.8347 & 0.8457 & 0.8380 & 0.8327 \\
		5     & 0.8542 & 0.8523 & 0.8571 & 0.8422 & 0.8410 & 0.8377 \\
		10    & 0.8550 & 0.8656 & \textbf{0.8679} & 0.8652 & 0.8642 & 0.8675 \\
		20    & 0.8555 & 0.8565 & 0.8590 & 0.8575 & 0.8586 & 0.8560 \\
		50    & 0.8357 & 0.8439 & 0.8410 & 0.8487 & 0.8444 & 0.8422 \\
		80    & 0.8371 & 0.8346 & 0.8392 & 0.8323 & 0.8197 & 0.8283 \\
		100   & 0.8345 & 0.8340 & 0.8240 & 0.8288 & 0.7964 & 0.8053 \\
		150   & 0.8134 & 0.8237 & 0.8158 & 0.8215 & 0.7589 & 0.7316 \\
		200   & 0.8222 & 0.8289 & 0.8058 & 0.8128 & 0.6626 & 0.6791 \\
		\hline
		\rowcolor[rgb]{ .906,  .902,  .902} \multicolumn{7}{|c|}{40 pix/class} \\

		1     & 0.8019 & 0.8289 & 0.8436 & 0.8488 & 0.8471 & 0.8431 \\
		5     & 0.8610 & 0.8712 & 0.8612 & 0.8589 & 0.8420 & 0.8483 \\
		10    & 0.8602 & 0.8728 & 0.8745 & \textbf{0.8760} & 0.8741 & 0.8744 \\
		20    & 0.8676 & 0.8675 & 0.8729 & 0.8665 & 0.8650 & 0.8698 \\
		50    & 0.8536 & 0.8624 & 0.8551 & 0.8600 & 0.8512 & 0.8537 \\
		80    & 0.8510 & 0.8568 & 0.8526 & 0.8510 & 0.8336 & 0.8382 \\
		100   & 0.8434 & 0.8472 & 0.8425 & 0.8449 & 0.8344 & 0.8422 \\
		150   & 0.8393 & 0.8320 & 0.8445 & 0.8172 & 0.7857 & 0.7957 \\
		200   & 0.8309 & 0.8264 & 0.8016 & 0.8217 & 0.7631 & 0.7681 \\
		\hline
	\end{tabular}%
	\label{salinas_rf_tuned_numeric_EMAP_NWFE_13_suppl}%
\end{table}%

\section{performance}
\begin{figure*}[t!]
	
	\begin{minipage}[b]{0.19\linewidth}
		\centering
		\centerline{\includegraphics[width=1\linewidth]{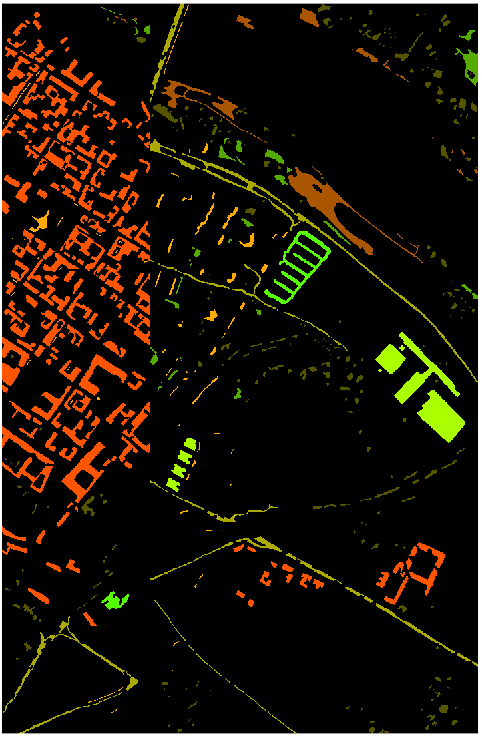}}
		\centerline{(a)}\medskip
	\end{minipage}
	\begin{minipage}[b]{0.19\linewidth}
		\centering
		\centerline{\includegraphics[width=1\linewidth]{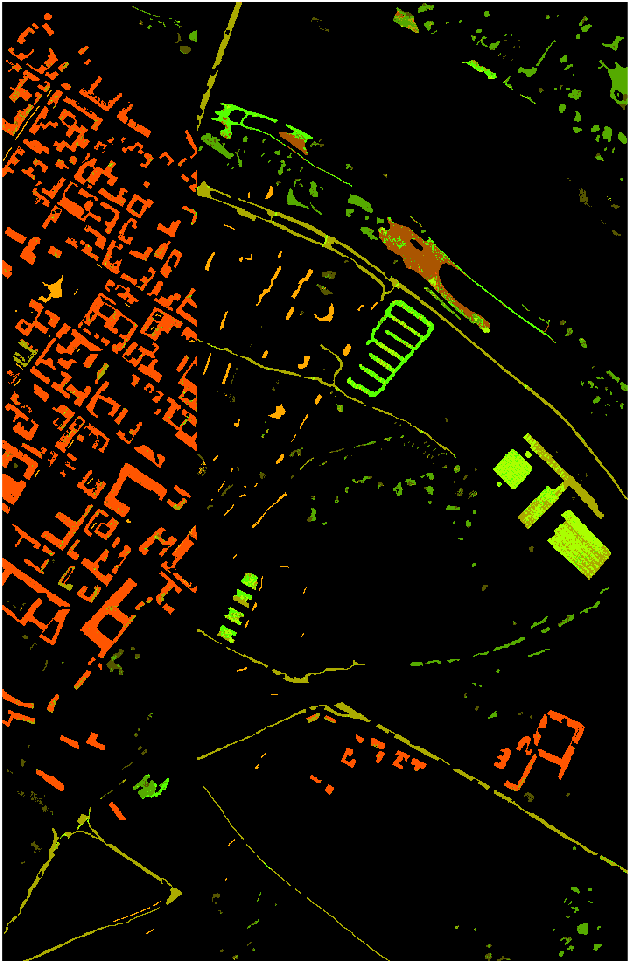}}
		\centerline{(b)}\medskip
	\end{minipage}
	\begin{minipage}[b]{0.19\linewidth}
		\centering
		\centerline{\includegraphics[width=1\linewidth]{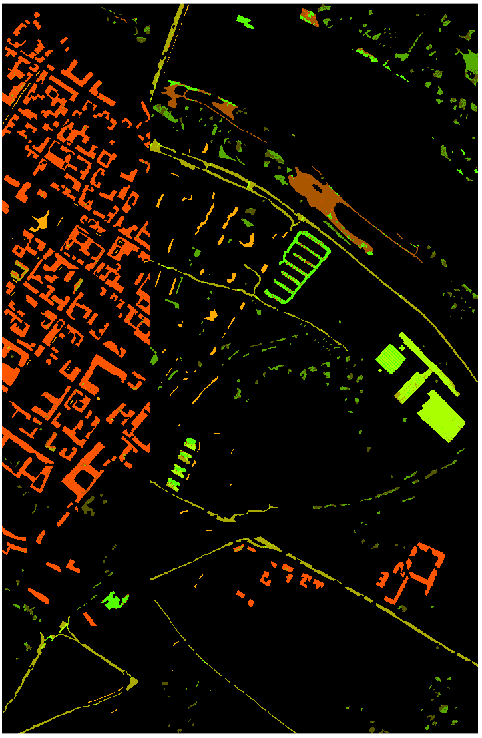}}
		\centerline{(c)}\medskip
	\end{minipage}
	\begin{minipage}[b]{0.19\linewidth}
		\centering
		\centerline{\includegraphics[width=1\linewidth]{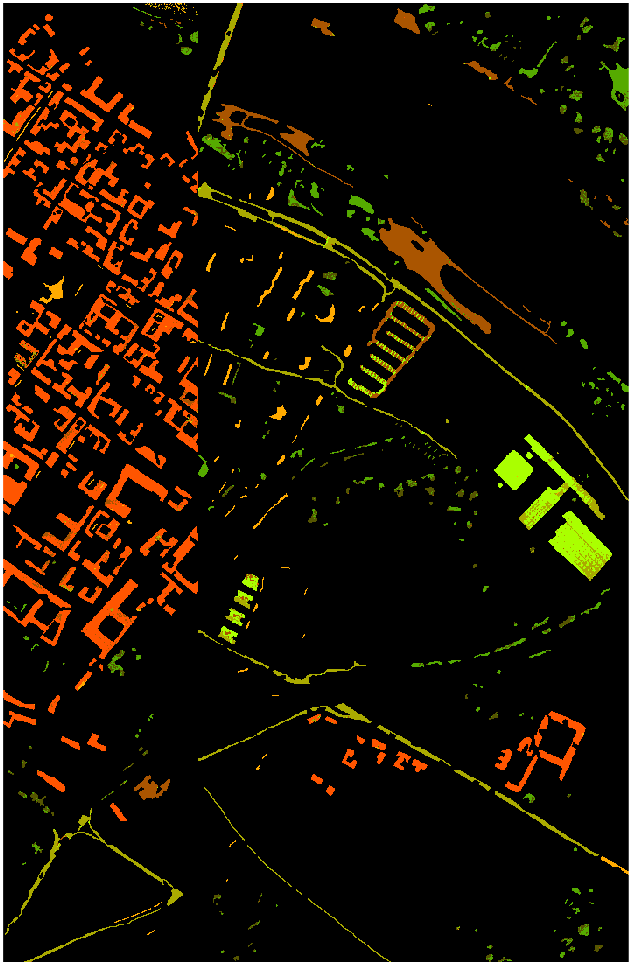}}
		\centerline{(d)}\medskip
	\end{minipage}
	\begin{minipage}[b]{0.19\linewidth}
		\centering
		\centerline{\includegraphics[width=1\linewidth]{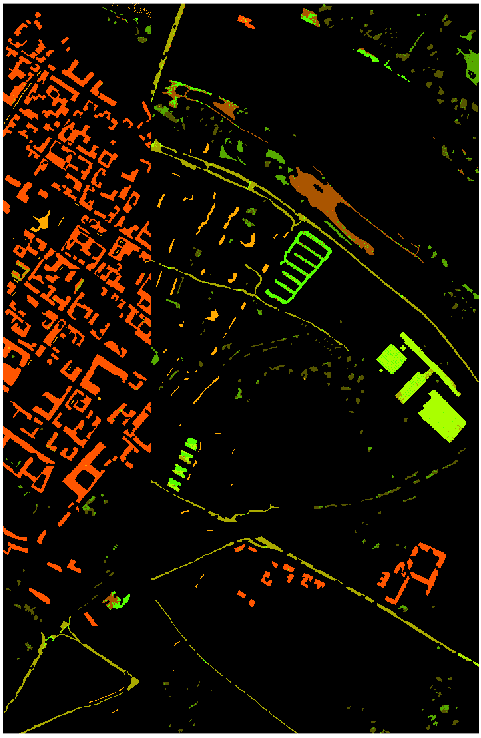}}
		\centerline{(e)}\medskip
	\end{minipage}
	\caption{Label maps on Pavia Centre using 13 training samples per class. (a) ground truth, (b) EMAP-PCA, (c) EMAP-PCA with $500$ synthetic samples per class, (d) EMAP-NWFE, (e) EMAP-NWFE with $500$ synthetic samples per class.}
	\label{Pavia_LM}
\end{figure*}

\begin{figure*}[t!]
	
	\begin{minipage}[b]{0.19\linewidth}
		\centering
		\centerline{\includegraphics[width=1\linewidth]{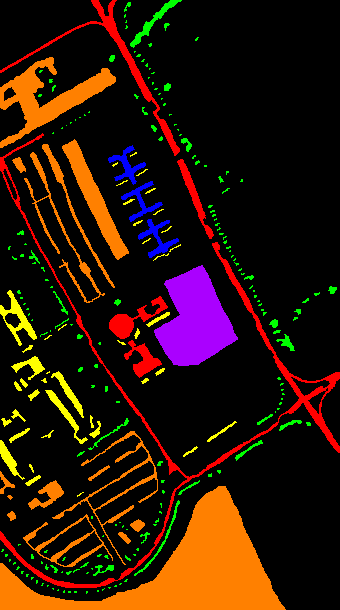}}
		\centerline{(a)}\medskip
	\end{minipage}
	\begin{minipage}[b]{0.19\linewidth}
		\centering
		\centerline{\includegraphics[width=1\linewidth]{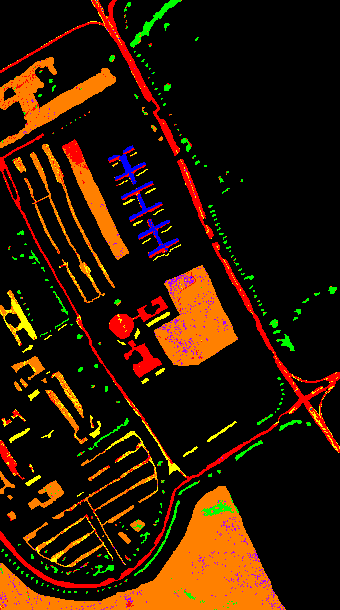}}
		\centerline{(b)}\medskip
	\end{minipage}
	\begin{minipage}[b]{0.19\linewidth}
		\centering
		\centerline{\includegraphics[width=1\linewidth]{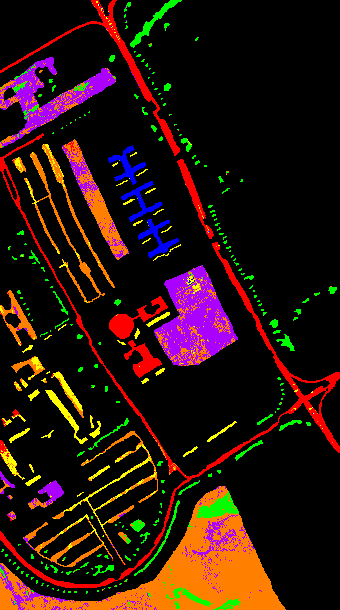}}
		\centerline{(c)}\medskip
	\end{minipage}
	\begin{minipage}[b]{0.19\linewidth}
		\centering
		\centerline{\includegraphics[width=1\linewidth]{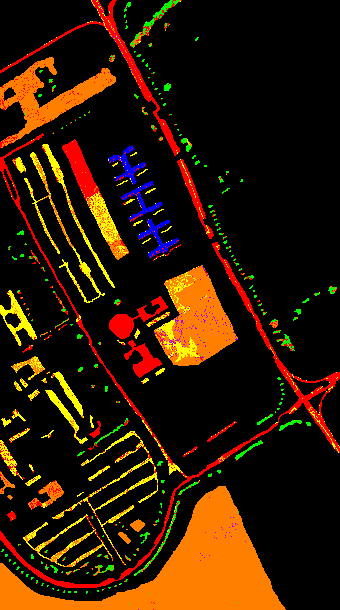}}
		\centerline{(d)}\medskip
	\end{minipage}
	\begin{minipage}[b]{0.19\linewidth}
		\centering
		\centerline{\includegraphics[width=1\linewidth]{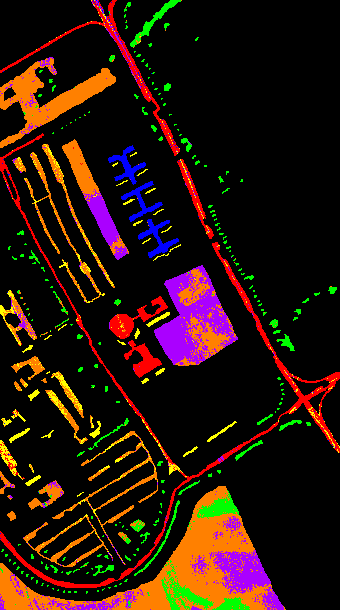}}
		\centerline{(e)}\medskip
	\end{minipage}
	\caption{Label maps on Pavia University using 13 training samples per class. (a) ground truth, (b) EMAP-PCA, (c) EMAP-PCA with $500$ synthetic samples per class, (d) EMAP-NWFE, (e) EMAP-NWFE with $500$ synthetic samples per class.}
	\label{PaviaU_LM}
\end{figure*}

Fig.~\ref{Pavia_LM} shows selected random forest label maps on Pavia Centre when adding synthetic samples. Analogously, Fig.~\ref{PaviaU_LM} shows selected random forest label maps on Pavia University when adding synthetic samples. Synthetic data augmentation improves the classification accuracy and avoids some misclassification. 

\clearpage

\clearpage\clearpage
\bibliographystyle{ieeetr}
\bibliography{ms}

\end{document}